\documentclass[10pt,twocolumn,letterpaper]{article}

\usepackage[pagenumbers]{cvpr}

\usepackage[hyphens]{url}

\usepackage{amsmath}
\usepackage{amssymb}

\usepackage{multirow}
\usepackage[table]{xcolor}
\usepackage{array}
\usepackage{tabularx}
\usepackage{booktabs}
\usepackage{makecell}

\newcolumntype{L}[1]{>{\raggedright\arraybackslash}p{#1}}
\newcolumntype{C}[1]{>{\centering\arraybackslash}p{#1}}
\newcolumntype{R}[1]{>{\raggedleft\arraybackslash}p{#1}}
\newcolumntype{Y}{>{\centering\arraybackslash}X}

\usepackage{algorithm}
\usepackage{algorithmic}

\usepackage{listings}
\usepackage{adjustbox} 
\usepackage{pifont}  
\definecolor{pacdpurple}{RGB}{239,234,248}

\definecolor{cvprblue}{rgb}{0.21,0.49,0.74}

\usepackage[pagebackref,breaklinks,colorlinks,allcolors=cvprblue]{hyperref}

\def\confName{CVPR}
\def\confYear{2027}

\title{PACD: Provenance-Aware Consensus Decoding for Mitigating Cross-Modal Hallucinations in Omnimodal Large Language Models}

\author{
\textbf{Chang Sun}$^{1}$, \textbf{Yijun Tian}$^{1*}$, \textbf{Zhenwei Tang}$^{2}$, \textbf{Yuchen Deng}$^{3,4}$,\\
\textbf{Yifan Simon Liu}$^{2}$, \textbf{Chenxu Wang}$^{1}$, \textbf{Wei Wang}$^{1}$\\[0.4em]
$^{1}$Xi'an Jiaotong University \qquad $^{2}$University of Toronto\\
$^{3}$Tsinghua University \qquad $^{4}$Pengcheng Laboratory\\[0.2em]
\href{https://github.com/SunChang726/PACD}{\textcolor{blue}{https://github.com/SunChang726/PACD}}
}

\usepackage{etoolbox}
\makeatletter
\patchcmd{\@maketitle}{\vskip .375in}{\vskip 0pt}{}{}
\patchcmd{\@maketitle}{\vspace*{24pt}}{\vspace*{13pt}}{}{}
\patchcmd{\@maketitle}{\vspace*{12pt}}{\vspace*{10pt}}{}{}
\patchcmd{\@maketitle}{\vskip .5em}{\vskip 0em}{}{}
\patchcmd{\abstract}{\vspace*{12pt}}{\vspace*{7pt}}{}{}
\patchcmd{\endabstract}{\vspace*{12pt}}{\vspace*{0pt}}{}{}
\patchcmd{\cvprsection}{-10pt plus -2pt minus -2pt}{-7pt plus -2pt minus -2pt}{}{}
\patchcmd{\cvprsection}{7pt}{4pt}{}{}
\makeatother

\begin{document}
\maketitle

\begingroup
\renewcommand{\thefootnote}{\fnsymbol{footnote}}
\footnotetext[1]{Corresponding author.}
\endgroup

\begin{abstract}
Omnimodal large language models (Omni-LLMs) integrate audio, video, and text for multimodal reasoning, yet remain vulnerable to cross-modal hallucinations, where one modality improperly influences predictions about another. Existing training-free decoding methods mitigate these errors through modality perturbation or relevance-based weighting. However, due to interference from irrelevant modality cues during fusion and difficulty detecting shared errors, these methods may suppress correct unimodal predictions and allow erroneous consensus to persist. To address these limitations, we propose Provenance-Aware Consensus Decoding (PACD), a training-free framework for mitigating cross-modal hallucinations. PACD comprises two complementary paths based on whether the branches reach consensus. For non-consensus cases, Relevance-Guided Candidate Arbitration identifies task-relevant unimodal evidence and selectively arbitrates between predictions using alternative candidates. For consensus cases, Provenance-Based Consensus Recovery leverages the outranking principle to evaluate the consistency and strength of layer-wise modality evidence for alternative answers, selectively resolving erroneous consensus under confidence constraints.
Extensive experiments on four Omni-LLMs and five benchmarks across two tasks demonstrate the superiority of PACD. Compared to existing methods, PACD achieves up to 27.6\% relative improvement over standard decoding across CMM, AVHBench, and SVHalluc for cross-modal hallucination mitigation, and an average relative improvement of 7.7\% over best baseline across OmniVideoBench and FutureOmni for general audio-visual reasoning.
\end{abstract}

\section{Introduction}

Omnimodal large language models (Omni-LLMs) have advanced multimodal perception and reasoning by integrating text, vision, and audio within a unified framework. Their ability to jointly process audio-visual information enables complex applications~\citep{LongVALE,VALOR}, including video understanding and interactive assistance. However, integrating multiple modalities does not necessarily guarantee reliable reasoning, as irrelevant information may interfere with task-relevant evidence and lead to inaccurate predictions~\citep{SECRET}.

Cross-modal hallucination remains a significant challenge for Omni-LLMs, where information from one modality improperly influences predictions from another~\citep{OmniHalluc-L}. For example, visual cues may cause a model to infer nonexistent sounds, while acoustic information may induce descriptions of visual events that never occurred~\citep{AVHBench}. Although multimodal fusion can exploit complementary evidence, it may also suppress reliable modality-specific information or reinforce incorrect predictions~\citep{CaRe,RED}. Improving cross-modal reliability therefore requires a better understanding of how different modalities contribute to the final answer.

\begin{figure}[t]
    \centering
    \includegraphics[width=1\linewidth]{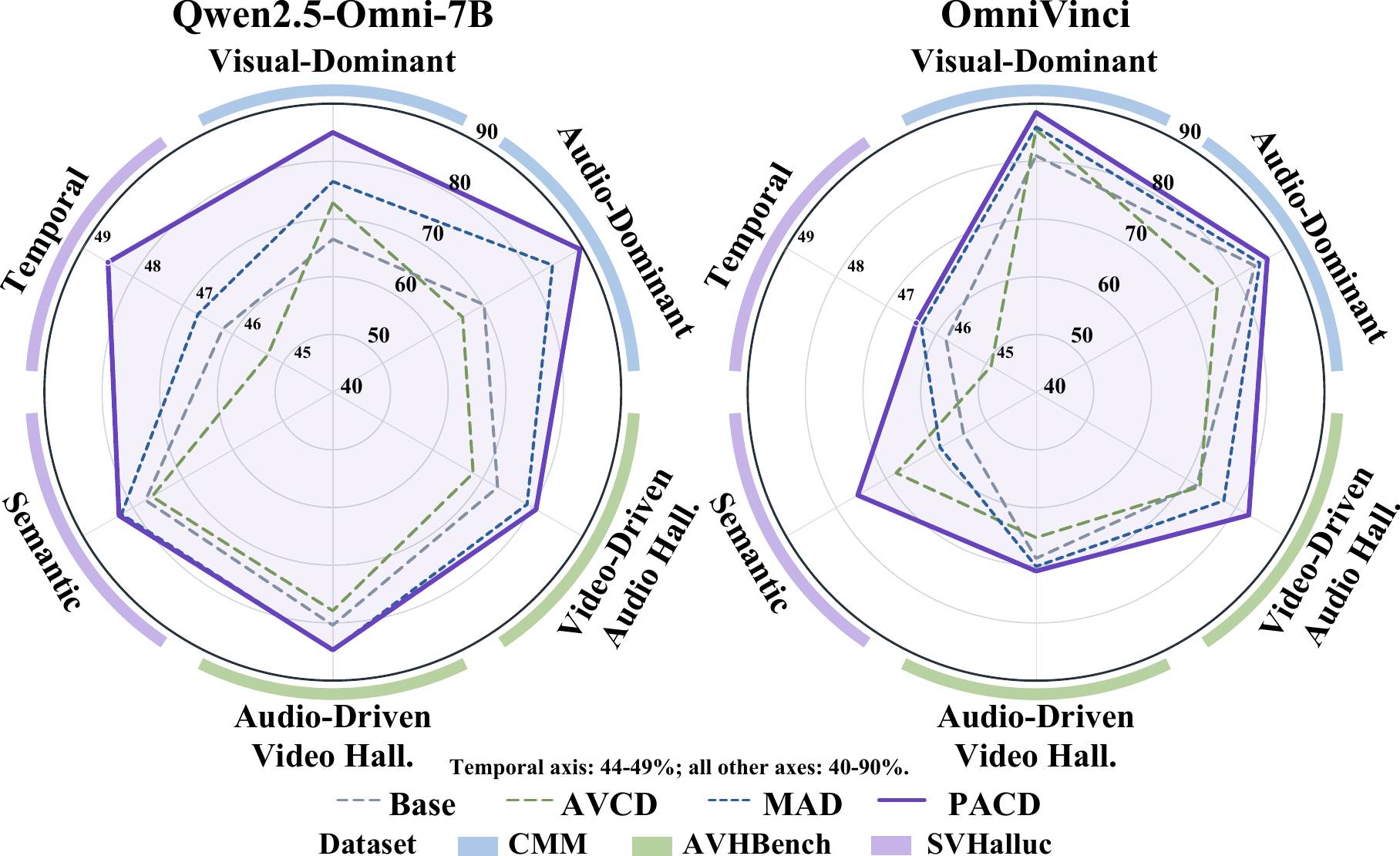}
    \caption{PACD achieves the state-of-the-art on CMM, AVHBench, and SVHalluc across Qwen2.5-Omni-7B and OmniVinci. }
    \label{fig:Intro}
\end{figure}

Recent training-free decoding methods attempt to mitigate cross-modal hallucinations by adjusting modality contributions during inference. For instance, AVCD~\citep{AVCD} employs attention-guided modality masking to construct contrastive predictions, while MAD~\citep{MAD} estimates query-specific modality relevance and combines predictions from different branches including audio-only, video-only, joint audio-visual, and text-only. To further investigate their limitations, we analyze OmniVinci on 2,405 questions from SVHalluc~\citep{SVHalluc}. Among 1,272 cases where the most relevant unimodal branch predicts correctly, MAD produces 111 incorrect ones. Moreover, among 1,791 cases where the audio-only, video-only, and joint audio-visual branches agree, 806 predictions are incorrect. MAD corrects only 23 of these cases, leaving 783 (97.1\%) unresolved.
This analysis reveals two critical challenges in mitigating cross-modal hallucinations: 
\begin{itemize}
    \item \textbf{C1:} Ignorance of reliable modality-specific predictions, where correct predictions from task-relevant unimodal branches can be suppressed during multimodal fusion.
    \item \textbf{C2:} Persistence of erroneous consensus, where  various branches can derive incorrect agreement and unfaithful predictions.
\end{itemize}

To address these challenges, we propose \textbf{Provenance-Aware Consensus Decoding (PACD)}, a training-free framework consisting of two complementary paths. For non-consensus cases, {Relevance Guided Candidate Arbitration (RGCA)} selects the prediction of the most relevant unimodal branch and constructs an alternative candidate using layer-wise modality contributions from the joint audio-visual branch. We then leverage the confidence of the routed candidate to preserve reliable evidence while avoiding unnecessary prediction replacement. For consensus cases, {Provenance Based Consensus Recovery (PBCR)} draws on the non-compensatory outranking principle~\cite{Outranking_Theory} to address potentially incorrect consensus by examining the consistency and strength of layer-wise evidence supporting alternative candidates. Combined with candidate uniqueness and confidence constraints, PBCR selectively revises the incorrect consensus predictions when sufficient evidence supports an alternative.

To evaluate the capability of PACD, we conduct extensive experiments on cross-modal hallucination mitigation and general audio-visual reasoning. The experimental results demonstrate that PACD consistently improves overall accuracy over baselines on CMM, AVHBench, SVHalluc, OmniVideoBench, and FutureOmni across the evaluated Omni-LLMs. As shown in Fig.~\ref{fig:Intro}, PACD achieves an average relative improvement of 14.2\% over standard decoding across CMM, AVHBench, and SVHalluc on Qwen2.5-Omni-7B, with relative gains of 6.1\%, 10.8\%, and 25.1\% on OmniVinci, respectively. Additional comparison experiments on generalization, ablation, efficiency and robustness further verify the effectiveness of PACD.

We summarize the contributions of this paper as follows:
\begin{itemize}
    \item We discover two critical limitations of existing methods in mitigating cross-modal hallucinations: the ignorance of reliable modality-specific predictions and the persistence of erroneous consensus.
    \item To address these challenges, we propose PACD, a training-free decoding framework with two complementary paths that improve the faithfulness of Omni-LLM predictions.
    \item We present RGCA and PBCR to address unreliable cross-modal predictions. RGCA combines relevance-guided candidate selection with provenance-informed confidence arbitration to preserve task-relevant unimodal evidence, while PBCR draws on the outranking principle to selectively resolve erroneous consensus through layer-wise evidence and confidence constraints.
    \item Extensive experiments show that PACD consistently outperforms existing baselines on CMM, AVHBench, and SVHalluc across four Omni-LLMs, with additional improvements on the general audio-visual benchmarks OmniVideoBench and FutureOmni.
\end{itemize}

\section{Related Work}
\textbf{Omnimodal Large Language Models.}
Multimodal large language models (MLLMs) have progressed from image-text understanding to complex audio-visual reasoning. Earlier MLLMs~\cite{Qwen3-VL,InternVL,VideoChat,Video-LLaVA,MedGemma,video-SALMONN} typically process different modalities through separate pathways. Recent omnimodal large language models (Omni-LLMs)~\cite{OneLLM,VITA,Baichuan-Omni,Qwen3-Omni,Qwen3.5-Omni}, including Qwen2.5-Omni and OmniVinci~\cite{Qwen2.5-Omni,OmniVinci}, integrate text, visual, and audio information within a unified framework, allowing temporal alignment and joint audio-visual reasoning~\cite{Ming-Omni,InteractiveOmni,Mini-Omni2}. However, unified multimodal processing does not always ensure reliable evidence integration. Irrelevant modality cues may interfere with task-relevant information, while agreement among modality-specific predictions does not necessarily guarantee correctness.

\noindent\textbf{Mitigating Hallucinations in Omni-LLMs.}
Hallucinations remain a major challenge for MLLMs, as generated responses may contradict multimodal evidence~\citep{POPE,HallusionBench,VideoHallucer}. Existing approaches include representation debiasing~\citep{HACL,Post-HocDebias}, architectural refinement~\citep{DeCo,RoPE,Nullu,OPERA}, and training-free decoding~\citep{CD,ED,HALC}. Training-free methods avoid model retraining and mitigate hallucinations through modality perturbations~\citep{VCD}, auxiliary descriptions~\citep{CODE}, instruction-level contrasts~\citep{ICD}, or layer-wise distribution differences~\citep{DoLa}. However, these methods primarily target vision-language models without considering audio-visual interactions.
Recent training-free methods extend hallucination mitigation to Omni-LLMs. AVCD~\citep{AVCD} constructs contrastive logits through attention-guided modality masking, while MAD~\citep{MAD} adaptively weights modality-specific predictions based on query relevance. However, relevance-guided fusion can suppress correct unimodal predictions, while agreement across branches may conceal shared errors. To address these issues, we propose PACD, which combines relevance-guided candidate arbitration with provenance-based consensus recovery to preserve reliable predictions and correct incorrect consensus.

\begin{figure*}[t]
    \centering
    \includegraphics[width=1\linewidth]{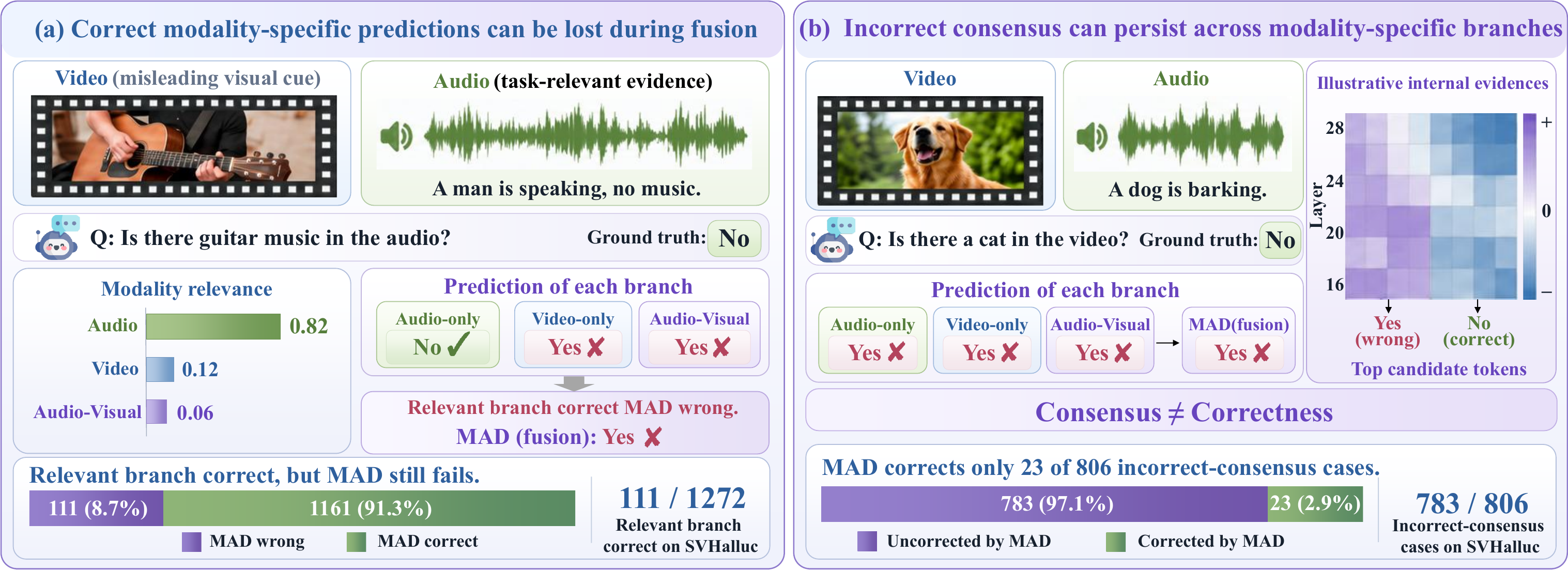}
    \caption{Motivation of PACD. (a) Correct modality-specific predictions may be lost during fusion. (b) Incorrect consensus among modality-specific branches often remains unresolved by MAD.}
    \label{fig:motivation}
\end{figure*}

\section{Preliminaries}
\label{sec:Preliminaries}

\subsection{Modality-Adaptive Decoding}

Modality-Adaptive Decoding (MAD) is a training-free decoding method that adjusts modality-specific contributions according to their relevance to the input query. Given audio $X_A$, video $X_V$, and textual query $X_Q$, MAD uses a fixed modality-query prompt $X_M$ to determine whether the query requires audio, video, or both. Let $u_A$, $u_V$, and $u_{AV}$ denote the logits corresponding to the responses ``audio'', ``video'', and ``both'', respectively. The modality-relevance scores are computed as
\begin{equation}
(r_A,r_V,r_{AV})
=
\operatorname{softmax}
\!\left([u_A,u_V,u_{AV}]\right).
\label{eq:mad_relevance}
\end{equation}

At decoding step $t$, let $y_{<t}$ denote previously generated tokens, $\mathcal V$ the vocabulary, and $\mathcal K=\{AV,A,V,T\}$ the four input configurations. These correspond to joint audio-visual input $X^{(AV)}=(X_A,X_V)$, audio-only input $X^{(A)}=X_A$, video-only input $X^{(V)}=X_V$, and text-only input $X^{(T)}=\varnothing$, respectively. All configurations retain the textual query $X_Q$. For each configuration $k$, the model produces next-token logits:
\begin{equation}
\begin{gathered}
z_k^t(v)
=
\operatorname{logit}_{\theta}
\!\left(
v\mid X^{(k)},X_Q,y_{<t}
\right),\\[-1mm]
k\in\mathcal K,
\qquad
v\in\mathcal V.
\end{gathered}
\label{eq:mad_branch_logits}
\end{equation}

MAD combines these logits using relevance-dependent weights:
\begin{equation}
\begin{aligned}
z_{\mathrm{MAD}}^t(v)
&=
\sum_{k\in\mathcal K}
w_k z_k^t(v),\\
w_{AV}
&=
2+2\gamma r_{AV},\\
w_A
&=
1-\gamma(r_{AV}-r_A),\\
w_V
&=
1-\gamma(r_{AV}-r_V),\\
w_T
&=
-\gamma(r_A+r_V).
\end{aligned}
\label{eq:mad_compact_decoding}
\end{equation}

Here, $\gamma>0$ controls the contrastive strength, and the weights satisfy $\sum_{k\in\mathcal K}w_k=4$. The audio-visual branch receives greater weight as $r_{AV}$ increases, while the unimodal weights depend on their relevance relative to the joint input. The negative text-only weight reduces the influence of language priors. The next token is selected by
\begin{equation}
\hat y_t
=
\arg\max_{v\in\mathcal V}
z_{\mathrm{MAD}}^t(v).
\label{eq:mad_next_token}
\end{equation}

By incorporating query-dependent modality relevance, MAD adaptively combines audio, visual, and joint audio-visual predictions to mitigate cross-modal hallucinations.

\begin{figure*}[t]
    \centering
    \includegraphics[width=1\linewidth]{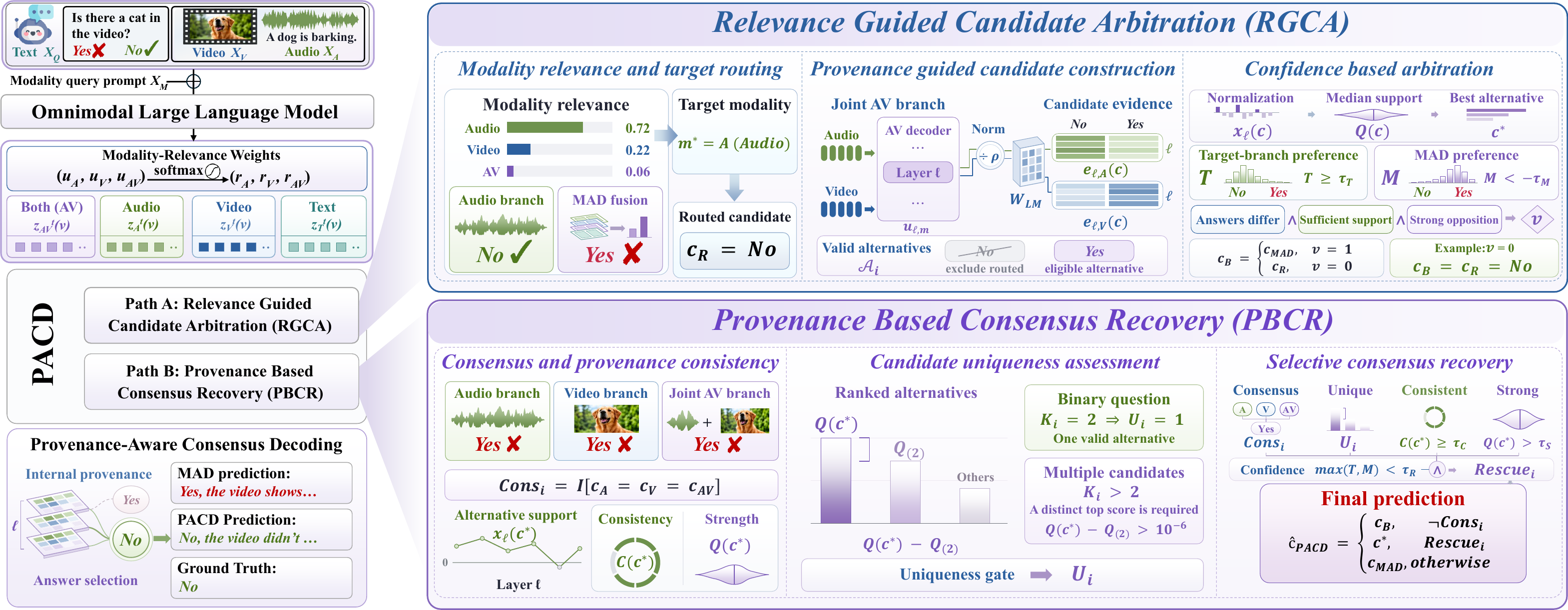}
    \caption{Overview of PACD. RGCA preserves reliable modality-specific predictions through relevance-guided routing and provenance-based arbitration. PBCR leverages layer-wise evidence to selectively correct incorrect consensus.}
    \label{fig:framework}
\end{figure*}

\subsection{Motivation}

MAD adaptively combines modality-specific predictions based on query relevance. However, relevance weighting does not always preserve correct unimodal predictions, and agreement across branches does not necessarily indicate correctness.

To investigate these issues, we analyze OmniVinci on 2,405 SVHalluc questions. As shown in Fig.~\ref{fig:motivation}(a), MAD produces 111 incorrect predictions among 1,272 cases where the most relevant unimodal branch predicts correctly. Figure~\ref{fig:motivation}(b) reveals another challenge: incorrect consensus. Among 1,791 cases where the audio-only ($A$), video-only ($V$), and audio-visual ($AV$) branches agree, 806 are incorrect. MAD corrects only 23 of these cases, leaving 783 (97.1\%) unresolved. Notably, even when all three branches predict the same incorrect answer, layer-wise contributions from the task-relevant modality can still favor the correct alternative, providing useful evidence for consensus correction.

These findings motivate PACD, which goes beyond modality relevance to examine candidate-level confidence and internal provenance evidence. PACD preserves reliable modality-specific predictions through relevance-guided arbitration and selectively corrects incorrect consensus using layer-wise modality contributions.

\section{Method}
\label{sec:method}

We propose \textbf{Provenance-Aware Consensus Decoding (PACD)}, a training-free framework that preserves reliable modality-specific predictions and selectively corrects incorrect consensus using internal provenance evidence. As illustrated in Fig.~\ref{fig:framework}, PACD follows two condition-specific paths depending on whether the audio-only, video-only, and joint audio-visual branches reach consensus.

For non-consensus cases, {Relevance Guided Candidate Arbitration (RGCA)} identifies the most relevant unimodal branch based on query-specific modality relevance. It constructs an alternative candidate from layer-wise provenance evidence and uses confidence-based arbitration to assess the predictive preferences of the target branch and MAD, preserving reliable modality-specific predictions while avoiding unnecessary replacement of fused predictions.

For consensus cases, Provenance Based Consensus Recovery (PBCR) is motivated by the non-compensatory outranking principle~\cite{Outranking_Theory}, which requires sufficient supporting evidence and the absence of strong counter-evidence before favoring an alternative. PBCR examines the shared prediction among the audio-only, video-only, and joint audio-visual branches, evaluates the consistency and strength of layer-wise provenance support for alternative candidates, and incorporates confidence constraints to selectively resolve erroneous consensus.

\subsection{Relevance Guided Candidate Arbitration}
\label{sec:routing_preservation}

For question $i$, let $\mathcal{C}_i=\{c_{i,1},\ldots,c_{i,K_i}\}$ be the valid answer candidates, where $K_i\geq 2$. Let $c_{\mathrm{MAD}}$, $c_A$, $c_V$, and $c_{AV}$ denote the MAD, audio-only, video-only, and joint audio-visual predictions, respectively.

\paragraph{Modality relevance and target routing.}
Given the relevance scores $(r_A,r_V,r_{AV})$ in Eq.~\eqref{eq:mad_relevance}, RGCA selects the target modality and routed candidate as
\begin{equation}
m^*=
\begin{cases}
A, & r_A>r_V,\\
V, & r_A\leq r_V,
\end{cases}
\qquad
c_R=c_{m^*}.
\label{eq:pacd_target_modality}
\end{equation}

\paragraph{Provenance guided candidate construction.}
To construct an alternative for confidence comparison, we examine modality-specific contributions within the joint audio-visual branch. Given $c_R$, the valid alternative set is $\mathcal{A}_i=\mathcal{C}_i\setminus\{c_R\}$.

At decoder layer $\ell$, let $\mathcal{S}_A$ and $\mathcal{S}_V$ denote the audio and visual token positions in the joint input. The attention weight from head $h$ to position $j$ at the answer-generation position is

\begin{equation}
a_{\ell,h,j}
=
\frac{
\exp\left(
\mathbf{q}_{\ell,h}^{\top}
\mathbf{k}_{\ell,h,j}/\sqrt{d_h}
\right)
}{
\sum_{j'\in\mathcal{P}}
\exp\left(
\mathbf{q}_{\ell,h}^{\top}
\mathbf{k}_{\ell,h,j'}/\sqrt{d_h}
\right)
},
\label{eq:pacd_attention_weight}
\end{equation}
where $\mathbf{q}_{\ell,h}$ and $\mathbf{k}_{\ell,h,j}$ are the query and key vectors, $d_h$ is the head dimension, and $\mathcal{P}$ denotes all valid attended positions.
We extract the attention contribution from modality $m\in\{A,V\}$ as
\begin{equation}
\mathbf{u}_{\ell,m}
=
W_{\ell}^{O}
\operatorname{Concat}_{h=1}^{H}
\left(
\sum_{j\in\mathcal{S}_m}
a_{\ell,h,j}\mathbf{v}_{\ell,h,j}
\right),
\label{eq:pacd_source_attention}
\end{equation}
where $\mathbf{v}_{\ell,h,j}$ is the value vector, $H$ is the number of attention heads, and $W_{\ell}^{O}$ is the attention output projection. Full-context attention normalization preserves the relative audio-visual weighting of these contributions.
For an RMSNorm-based output layer, we project each contribution into candidate-logit space as
\begin{equation}
e_{\ell,m}(c)
=
W_{\mathrm{LM}}[\phi_i(c)]^\top
\left(
\frac{
\mathbf{g}\odot\mathbf{u}_{\ell,m}
}{
\rho
}
\right),
\label{eq:pacd_provenance_contribution}
\end{equation}
where $W_{\mathrm{LM}}[\phi_i(c)]$ is the output projection vector for candidate $c$, $\mathbf{g}$ is the RMSNorm scale, and $\odot$ denotes element-wise multiplication. The normalization denominator is
\begin{equation}
\rho=
\sqrt{
\frac{1}{D}
\left\|
\mathbf{h}_{\mathrm{pre}}
\right\|_2^2
+\epsilon_{\mathrm{norm}}
},
\label{eq:pacd_provenance_norm}
\end{equation}
where $\mathbf{h}_{\mathrm{pre}}$ is the hidden state before output normalization and $D$ is its dimension. For other normalization architectures, we use the corresponding model-specific projection.

The resulting $e_{\ell,m}(c)$ measures the modality-specific attention contribution in candidate-logit space under a fixed normalization scale. These signals are extracted from the existing joint forward pass without modifying attention outputs or hidden states, serving as internal provenance indicators rather than exact causal attributions.

\paragraph{Confidence based arbitration.}
For each $c\in\mathcal{A}_i$, RGCA computes the layer-wise provenance difference from $c_R$:
\begin{equation}
d_{\ell,m}(c)=e_{\ell,m}(c)-e_{\ell,m}(c_R),
\label{eq:pacd_source_difference}
\end{equation}
where positive values indicate greater support for $c$ than for $c_R$ from modality $m$.
To account for variations in contribution magnitude, we define a candidate-specific scale over the last $N$ recorded layers $\mathcal{L}_N$:
\begin{equation}
\begin{aligned}
s_e(c)=\max\Bigl\{&
\operatorname{Quantile}_{0.75}
\bigl(\{|d_{\ell,m}(c)|:
\\[-2pt]
&\ell\in\mathcal{L}_N,\,
m\in\{A,V\}\}\bigr),
10^{-5}
\Bigr\}.
\end{aligned}
\label{eq:pacd_source_scale}
\end{equation}

For the selected modality $m^*$, the normalized contribution difference is
\begin{equation}
x_\ell(c)=\frac{d_{\ell,m^*}(c)}{s_e(c)}.
\label{eq:pacd_normalized_source}
\end{equation}

We select the strongest alternative based on median layer-wise evidence:
\begin{equation}
c^*=\arg\max_{c\in\mathcal{A}_i}Q(c),
\quad
Q(c)=\operatorname{median}_{\ell\in\mathcal{L}_N}x_\ell(c).
\label{eq:pacd_alternative}
\end{equation}

Therefore, $c^*$ is determined by the target modality's internal provenance evidence without using ground-truth answers. RGCA then compares $c_R$ and $c^*$ using the target branch and MAD predictions. Following Eq.~\eqref{eq:mad_branch_logits}, the logit for candidate $c$ is $z_b^t(c)=z_b^t\bigl(\phi_i(c)\bigr)$, where $b\in\mathcal{B}$ and $\mathcal{B}=\{\mathrm{MAD},AV,A,V,T\}$. MAD logits are obtained from Eq.~\eqref{eq:mad_compact_decoding}. To normalize candidate-level logit differences across branches, we define
\begin{equation}
s_z=\max\left\{\operatorname{median}_{b\in\mathcal{B}}\left|z_b^t(c_R)-z_b^t(c^*)\right|,\,0.5\right\}.
\label{eq:pacd_logit_scale}
\end{equation}

The median yields a robust scale, with a lower bound preventing amplification of small differences. We compute two normalized confidence margins:

\begin{equation}
\begin{aligned}
T &= \frac{z_{m^*}^t(c_R)-z_{m^*}^t(c^*)}{s_z},\\
M &= \frac{z_{\mathrm{MAD}}^t(c_R)-z_{\mathrm{MAD}}^t(c^*)}{s_z}.
\end{aligned}
\label{eq:pacd_confidence_margins}
\end{equation}

Here, $T$ and $M$ measure the preferences of the target branch and MAD for $c_R$ over $c^*$, respectively. Positive values favor $c_R$, while negative values favor $c^*$. RGCA combines these margins through a confidence veto:
\begin{equation}
\begin{aligned}
v={}&\mathbb{I}[c_R\neq c_{\mathrm{MAD}}]\land\mathbb{I}[T<\tau_T]\land\mathbb{I}[M<-\tau_M],
\end{aligned}
\label{eq:pacd_veto}
\end{equation}

where $\tau_T$ and $\tau_M$ are the confidence thresholds for the target branch and MAD, respectively. RGCA outputs $c_B=c_{\mathrm{MAD}}$ when $v=1$ and $c_B=c_R$ otherwise, preserving task-relevant predictions while considering fused confidence.

\subsection{Provenance Based Consensus Recovery}
\label{sec:consensus_recovery}

For consensus cases, PBCR uses provenance consistency and predictive confidence to assess whether $c^*$ should replace the shared prediction.

\paragraph{Consensus and provenance consistency.}
We first identify consensus among the audio-only, video-only, and joint audio-visual branches as $\operatorname{Cons}_i=\mathbb{I}[c_A=c_V=c_{AV}]$.

To assess internal support for the alternative, PBCR uses the normalized layer-wise contributions $x_\ell(c^*)$ from Eq.~\eqref{eq:pacd_normalized_source}. We define the provenance consistency score as the proportion of examined layers in which the target modality favors $c^*$ over $c_R$:

\begin{equation}
C(c^*)=\frac{1}{|\mathcal{L}_N|}\sum_{\ell\in\mathcal{L}_N}\mathbb{I}[x_\ell(c^*)>0].
\label{eq:pacd_source_consistency}
\end{equation}
Here, each examined decoder layer provides a criterion for comparing $c^*$ against $c_R$. Accordingly, $C(c^*)\in[0,1]$ serves as an unweighted concordance index, measuring the proportion of layers supporting the alternative, while $Q(c^*)$ from Eq.~\eqref{eq:pacd_alternative} captures the median strength of this support.

\begin{table*}[t]
    \centering
    \caption{
        Main results on cross-modal hallucination benchmarks.
        We compare the base decoding strategy (Base), AVCD, MAD,
        and our proposed PACD.
        Results are reported in accuracy (\%).
        Arrows indicate relative gains over Base.
    }
    \label{tab:main_results}
    \small
    \setlength{\tabcolsep}{0.5mm}
    \renewcommand{\arraystretch}{0.92}

    \begin{tabular*}{\textwidth}{
        @{\extracolsep{\fill}}
        p{1.30cm}
        C{1.70cm}
        C{1.70cm}
        C{1.70cm}
        C{2.05cm}
        C{2.05cm}
        C{2.05cm}
        C{1.05cm}
        C{1.05cm}
        C{2.10cm}
        @{}
    }
        \toprule

        \multirow[c]{2}{*}{\raisebox{-0.98em}[0pt][0pt]{\textbf{Method}}}
        & \multicolumn{3}{c}{\textbf{CMM}}
        & \multicolumn{3}{c}{\textbf{AVHBench}}
        & \multicolumn{3}{c}{\textbf{SVHalluc}} \\

        \cmidrule(lr){2-4}
        \cmidrule(lr){5-7}
        \cmidrule(lr){8-10}

        &
        \parbox[c][2.6em][c]{1.70cm}{
            \centering\textbf{Visual Dom.}
        }
        &
        \parbox[c][2.6em][c]{1.70cm}{
            \centering\textbf{Audio Dom.}
        }
        &
        \parbox[c][2.6em][c]{1.70cm}{
            \centering\textbf{Overall Acc.}
        }
        &
        \parbox[c][2.6em][c]{2.05cm}{
            \centering
            \textbf{Video-Driven}\\
            \textbf{Audio Hall.}
        }
        &
        \parbox[c][2.6em][c]{2.05cm}{
            \centering
            \textbf{Audio-Driven}\\
            \textbf{Video Hall.}
        }
        &
        \parbox[c][2.6em][c]{2.05cm}{
            \centering\textbf{Overall Acc.}
        }
        &
        \parbox[c][2.6em][c]{1.05cm}{
            \centering\textbf{Sem.}
        }
        &
        \parbox[c][2.6em][c]{1.05cm}{
            \centering\textbf{Temp.}
        }
        &
        \parbox[c][2.6em][c]{2.10cm}{
            \centering\mbox{\textbf{Overall Acc.}}
        }
        \\

        \midrule

        \multicolumn{10}{c}{\textbf{VideoLLaMA2-AV-7B}} \\

        \midrule

        Base
        & 72.3 & 82.3 & 77.3
        & 70.8 & 76.1 & 72.6
        & 57.2 & 43.8 & 51.7 \\

        AVCD
        & 69.8 & \textbf{84.8} & 77.3
        & 78.0 & \textbf{80.1} & 78.7
        & 62.5 & \textbf{45.2} & 55.4 \\

        MAD
        & 81.3 & 80.5 & 80.9
        & 79.7 & 77.0 & 78.8
        & 65.1 & 44.5 & 56.7 \\

        \rowcolor{pacdpurple}
        \textbf{PACD}
        & \textbf{86.5} & 78.3
        & \textbf{82.4\,{\scriptsize($\uparrow$6.6\%)}}
        & \textbf{81.2} & 79.4
        & \textbf{80.6\,{\scriptsize($\uparrow$11.0\%)}}
        & \textbf{71.0} & 45.0
        & \textbf{60.4\,{\scriptsize($\uparrow$16.8\%)}} \\

        \midrule

        \multicolumn{10}{c}{\textbf{OmniVinci}} \\

        \midrule

        Base
        & 81.0 & 83.8 & 82.4
        & 72.2 & 68.8 & 71.1
        & 54.5 & 45.8 & 50.9 \\

        AVCD
        & 85.5 & 76.3 & 80.9
        & 72.8 & 65.2 & 70.3
        & 68.0 & 44.9 & 58.5 \\

        MAD
        & 86.0 & 84.8 & 85.4
        & 77.6 & 70.2 & 75.1
        & 59.2 & 46.3 & 53.9 \\

        \rowcolor{pacdpurple}
        \textbf{PACD}
        & \textbf{88.5} & \textbf{86.3}
        & \textbf{87.4\,{\scriptsize($\uparrow$6.1\%)}}
        & \textbf{82.6} & \textbf{71.0}
        & \textbf{78.8\,{\scriptsize($\uparrow$10.8\%)}}
        & \textbf{75.7} & \textbf{46.4}
        & \textbf{63.7\,{\scriptsize($\uparrow$25.1\%)}} \\

        \midrule

        \multicolumn{10}{c}{\textbf{Qwen2.5-Omni-3B}} \\

        \midrule

        Base
        & 70.8 & 85.3 & 78.0
        & 73.4 & 80.5 & 75.8
        & 72.7 & 42.1 & 60.2 \\

        AVCD
        & 75.5 & 67.0 & 71.3
        & 70.5 & 77.1 & 72.7
        & 66.2 & 45.3 & 57.7 \\

        MAD
        & 74.5 & 86.3 & 80.4
        & 80.1 & \textbf{82.5} & 80.9
        & 77.9 & 45.2 & 64.5 \\

        \rowcolor{pacdpurple}
        \textbf{PACD}
        & \textbf{80.8} & \textbf{88.5}
        & \textbf{84.6\,{\scriptsize($\uparrow$8.5\%)}}
        & \textbf{82.7} & 81.8
        & \textbf{82.4\,{\scriptsize($\uparrow$8.7\%)}}
        & \textbf{78.0} & \textbf{47.6}
        & \textbf{65.6\,{\scriptsize($\uparrow$9.0\%)}} \\

        \midrule

        \multicolumn{10}{c}{\textbf{Qwen2.5-Omni-7B}} \\

        \midrule

        Base
        & 66.5 & 70.3 & 68.4
        & 73.0 & 80.4 & 75.5
        & 77.5 & 46.2 & 64.7 \\

        AVCD
        & 72.8 & 66.0 & 69.4
        & 68.1 & 77.9 & 71.3
        & 76.2 & 45.3 & 63.5 \\

        MAD
        & 76.5 & 84.0 & 80.3
        & 78.9 & \textbf{84.7} & 80.8
        & 82.3 & 46.7 & 67.8 \\

        \rowcolor{pacdpurple}
        \textbf{PACD}
        & \textbf{85.0} & \textbf{89.5}
        & \textbf{87.3\,{\scriptsize($\uparrow$27.6\%)}}
        & \textbf{80.7} & \textbf{84.7}
        & \textbf{82.0\,{\scriptsize($\uparrow$8.6\%)}}
        & \textbf{82.8} & \textbf{48.5}
        & \textbf{68.8\,{\scriptsize($\uparrow$6.3\%)}} \\

        \bottomrule
    \end{tabular*}
\end{table*}

\paragraph{Candidate uniqueness assessment.}
PBCR checks whether $c^*$ is distinctly supported over other alternatives. Let $Q_{(2)}$ denote the second-highest provenance score in $\mathcal{A}_i$. Candidate uniqueness is defined as
\begin{equation}
U_i=
\begin{cases}
1, & K_i=2,\\
\mathbb{I}[Q(c^*)-Q_{(2)}>10^{-6}], & K_i>2.
\end{cases}
\label{eq:pacd_unique_challenger}
\end{equation}
Binary questions have a unique alternative; for $K_i>2$, recovery requires the specified provenance margin.


\paragraph{Selective consensus recovery.} Following the outranking principle, PBCR requires both sufficient concordant support and the absence of strong opposing evidence before revising a consensus prediction. Specifically, it combines provenance consistency and strength, candidate uniqueness, and the confidence margins $T$ and $M$ from RGCA through a non-compensatory decision rule. The recovery condition is defined as

\begin{equation}
\begin{aligned}
\operatorname{Rescue}_i={}&\operatorname{Cons}_i\land U_i\land\mathbb{I}[C(c^*)\geq\tau_C]\land\\
&\mathbb{I}[Q(c^*)>\tau_S]\land\mathbb{I}[\max(T,M)<\tau_R].
\end{aligned}
\label{eq:pacd_rescue_gate}
\end{equation}
Here, $\tau_C$ and $\tau_S$ specify the required concordance and strength thresholds for provenance support, respectively. The maximum confidence margin $\max(T,M)$ captures the largest branch-level margin favoring the routed candidate over the alternative, and its comparison with $\tau_R$ acts as a veto-like constraint. Together with candidate uniqueness, these non-compensatory conditions restrict consensus revision to alternatives satisfying the specified provenance consistency, evidence strength, candidate uniqueness, and confidence constraints.
Finally, PACD combines the RGCA arbitration result $c_B$ with the PBCR recovery decision:

\begin{equation}
\hat{c}_{\mathrm{PACD}}=
\begin{cases}
c_B, & \neg\operatorname{Cons}_i,\\
c^*, & \operatorname{Rescue}_i,\\
c_{\mathrm{MAD}}, & \text{otherwise}.
\end{cases}
\label{eq:pacd_final_prediction}
\end{equation}
Non-consensus cases use the RGCA prediction $c_B$; consensus cases adopt $c^*$ only when the recovery conditions hold, otherwise retaining MAD.

\begin{table*}[t]
    \centering
    \caption{Comparison of decoding strategies on CMM, AVHBench, and SVHalluc using OmniVinci.}
    \label{tab:fusion_ablation}
    \small
    \setlength{\tabcolsep}{0.5mm}
    \renewcommand{\arraystretch}{0.92}

    \begin{tabular*}{\textwidth}{
        @{\extracolsep{\fill}}
        p{1.30cm}
        C{1.70cm}
        C{1.70cm}
        C{1.70cm}
        C{2.05cm}
        C{2.05cm}
        C{2.05cm}
        C{1.05cm}
        C{1.05cm}
        C{2.10cm}
        @{}
    }
        \toprule

        \multirow[c]{2}{*}{\raisebox{-0.98em}[0pt][0pt]{\textbf{Method}}}
        & \multicolumn{3}{c}{\textbf{CMM}}
        & \multicolumn{3}{c}{\textbf{AVHBench}}
        & \multicolumn{3}{c}{\textbf{SVHalluc}} \\

        \cmidrule(lr){2-4}
        \cmidrule(lr){5-7}
        \cmidrule(lr){8-10}

        &
        \parbox[c][2.6em][c]{1.70cm}{
            \centering\textbf{Visual Dom.}
        }
        &
        \parbox[c][2.6em][c]{1.70cm}{
            \centering\textbf{Audio Dom.}
        }
        &
        \parbox[c][2.6em][c]{1.70cm}{
            \centering\textbf{Overall Acc.}
        }
        &
        \parbox[c][2.6em][c]{2.05cm}{
            \centering
            \textbf{Video-Driven}\\
            \textbf{Audio Hall.}
        }
        &
        \parbox[c][2.6em][c]{2.05cm}{
            \centering
            \textbf{Audio-Driven}\\
            \textbf{Video Hall.}
        }
        &
        \parbox[c][2.6em][c]{2.05cm}{
            \centering\textbf{Overall Acc.}
        }
        &
        \parbox[c][2.6em][c]{1.05cm}{
            \centering\textbf{Sem.}
        }
        &
        \parbox[c][2.6em][c]{1.05cm}{
            \centering\textbf{Temp.}
        }
        &
        \parbox[c][2.6em][c]{2.10cm}{
            \centering\mbox{\textbf{Overall Acc.}}
        }
        \\

        \midrule

        Base
        & 81.0 & 83.8 & 82.4
        & 72.2 & 68.8 & 71.1
        & 54.5 & 45.8 & 50.9 \\

        Uniform
        & 86.0 & 85.8 & 85.9
        & 76.2 & 68.6 & 73.7
        & 57.0 & 46.2 & 52.6 \\

        Argmax
        & 86.3 & 82.3 & 84.3
        & 74.7 & 68.8 & 72.7
        & 59.6 & 45.1 & 53.6 \\

        MAD
        & 86.0 & 84.8 & 85.4
        & 77.6 & 70.2 & 75.1
        & 59.2 & 46.3 & 53.9 \\

        \rowcolor{pacdpurple}
        \textbf{PACD}
        & \textbf{88.5} & \textbf{86.3} & \textbf{87.4}
        & \textbf{82.6} & \textbf{71.0} & \textbf{78.8}
        & \textbf{75.7} & \textbf{46.4} & \textbf{63.7} \\

        \bottomrule
    \end{tabular*}
\end{table*}

\begin{table*}[t]
    \centering
    \caption{Component ablation of PACD on CMM, AVHBench, and SVHalluc using OmniVinci.}
    \label{tab:components_ablation}
    \small
    \setlength{\tabcolsep}{0.5mm}
    \renewcommand{\arraystretch}{0.92}

    \begin{tabular*}{\textwidth}{
        @{\extracolsep{\fill}}
        p{1.30cm}
        C{1.70cm}
        C{1.70cm}
        C{1.70cm}
        C{2.05cm}
        C{2.05cm}
        C{2.05cm}
        C{1.05cm}
        C{1.05cm}
        C{2.10cm}
        @{}
    }
        \toprule

        \multirow[c]{2}{*}{\raisebox{-0.98em}[0pt][0pt]{\textbf{Method}}}
        & \multicolumn{3}{c}{\textbf{CMM}}
        & \multicolumn{3}{c}{\textbf{AVHBench}}
        & \multicolumn{3}{c}{\textbf{SVHalluc}} \\

        \cmidrule(lr){2-4}
        \cmidrule(lr){5-7}
        \cmidrule(lr){8-10}

        &
        \parbox[c][2.6em][c]{1.70cm}{\centering\textbf{Visual Dom.}}
        &
        \parbox[c][2.6em][c]{1.70cm}{\centering\textbf{Audio Dom.}}
        &
        \parbox[c][2.6em][c]{1.70cm}{\centering\textbf{Overall Acc.}}
        &
        \parbox[c][2.6em][c]{2.05cm}{\centering\textbf{Video-Driven}\\\textbf{Audio Hall.}}
        &
        \parbox[c][2.6em][c]{2.05cm}{\centering\textbf{Audio-Driven}\\\textbf{Video Hall.}}
        &
        \parbox[c][2.6em][c]{2.05cm}{\centering\textbf{Overall Acc.}}
        &
        \parbox[c][2.6em][c]{1.05cm}{\centering\textbf{Sem.}}
        &
        \parbox[c][2.6em][c]{1.05cm}{\centering\textbf{Temp.}}
        &
        \parbox[c][2.6em][c]{2.10cm}{\centering\mbox{\textbf{Overall Acc.}}}
        \\

        \midrule

        Base
        & 81.0 & 83.8 & 82.4
        & 72.2 & 68.8 & 71.1
        & 54.5 & 45.8 & 50.9 \\

        Routing
        & 87.8 & 79.3 & 83.5
        & 77.7 & 67.6 & 74.3
        & 59.7 & 44.9 & 53.6 \\

        RGCA
        & 88.3 & 85.5 & 86.9
        & 81.6 & 69.8 & 77.7
        & 59.8 & 45.5 & 54.0 \\

        R+PBCR
        & 88.0 & 80.0 & 84.0
        & 78.7 & 69.3 & 75.6
        & 75.0 & 45.4 & 62.9 \\

        \rowcolor{pacdpurple}
        \textbf{PACD}
        & \textbf{88.5} & \textbf{86.3} & \textbf{87.4}
        & \textbf{82.6} & \textbf{71.0} & \textbf{78.8}
        & \textbf{75.7} & \textbf{46.4} & \textbf{63.7} \\

        \bottomrule
    \end{tabular*}
\end{table*}

\section{Experiments}
\subsection{Experimental Setting}

\paragraph{Datasets and Methods.}
We evaluate PACD on five benchmarks. AVHBench~\cite{AVHBench} and CMM~\cite{CMM} assess cross-modal hallucinations, focusing on audio-visual interference and modality dominance. SVHalluc~\cite{SVHalluc} evaluates semantic and temporal hallucinations in speech-visual understanding. OmniVideoBench~\cite{OmniVideoBench} and FutureOmni~\cite{FutureOmni} assess general audio-visual reasoning. Together, these benchmarks provide a comprehensive evaluation of hallucination mitigation and general reasoning.
We compare PACD with two training-free decoding methods that have the best performance: AVCD~\cite{AVCD} and MAD~\cite{MAD}. AVCD uses attention-guided modality masking to construct contrastive logits for hallucination mitigation, while MAD adaptively combines modality-specific predictions using query-specific relevance scores.

\paragraph{Implementation Details.}
We evaluate PACD on four Omni-LLMs: VideoLLaMA2-AV~\cite{VideoLLaMA2-AV}, OmniVinci~\cite{OmniVinci}, Qwen2.5-Omni-3B, and Qwen2.5-Omni-7B~\cite{Qwen2.5-Omni}. All experiments are conducted on NVIDIA A100 GPUs with 80 GB of memory. Following prior work, videos are sampled at 2 fps, with a maximum of 768 frames per video. The contrastive strength parameter is fixed at $\gamma=2.5$, while the hyperparameter sensitivity analysis of PACD is detailed in Section 9 of the Supplementary Material. Code is available in the Supplementary Material as well.

\subsection{Performance Comparison}
We evaluate PACD on three cross-modal hallucination benchmarks: CMM, AVHBench, and SVHalluc. As shown in Table~\ref{tab:main_results}, PACD achieves the highest overall accuracy across all 12 model--benchmark combinations, outperforming the evaluated baselines with improvements of up to 9.8 percentage points over MAD. Specifically, on CMM, PACD reaches 87.3\% overall accuracy with Qwen2.5-Omni-7B, exceeding Base and MAD by 18.9 and 7.0 percentage points, respectively. On AVHBench, PACD improves overall accuracy by 6.5--8.0 percentage points over Base across the four backbones. On SVHalluc, PACD increases semantic and overall accuracy by 21.2 and 12.8 percentage points over Base on OmniVinci, respectively. These results demonstrate PACD's effectiveness in mitigating diverse forms of cross-modal hallucination across different Omni-LLMs and improving the reliability of audio-visual reasoning.

\subsection{Ablation Study}
Due to space limitations, we present the ablation results on CMM, AVHBench, and SVHalluc using OmniVinci in the main paper. Results for the other three backbones are provided in the Supplementary Material.

\paragraph{Comparison with Relevance Weighting Strategies.}
We compare PACD with two simplified weighting strategies. Uniform assigns equal relevance to all modalities ($r_A=r_V=r_{AV}=1/3$), while Argmax converts the relevance distribution into a one-hot assignment. Both apply the resulting MAD decoding weights without candidate arbitration or consensus recovery. As shown in Table~\ref{tab:fusion_ablation}, PACD outperforms both strategies by 1.5, 5.1, and 10.1 percentage points over the stronger baseline on CMM, AVHBench, and SVHalluc, respectively. These results highlight the benefits of combining relevance-aware decoding, confidence-based arbitration, and provenance-based consensus recovery.

\paragraph{Contribution of RGCA and PBCR.}
To assess the contributions of RGCA and PBCR, we compare PACD with three simplified variants. Routing adopts the prediction of the most relevant unimodal branch. RGCA adds provenance-guided candidate construction and confidence-based arbitration, while Routing + PBCR performs consensus recovery without arbitration. As shown in Table~\ref{tab:components_ablation}, RGCA improves overall accuracy over Routing by 3.4 percentage points on both CMM and AVHBench. Routing + PBCR improves overall accuracy by 9.3 percentage points on SVHalluc, increasing semantic accuracy from 59.7\% to 75.0\%. With both paths enabled, PACD achieves overall accuracies of 87.4\%, 78.8\%, and 63.7\% on the three benchmarks, respectively. These results highlight the complementary roles of RGCA in candidate arbitration and PBCR in correcting erroneous consensus.

\begin{table}[t]
    \centering
    \caption{Generalization performance on OmniVideoBench and FutureOmni. Arrows indicate relative gains over MAD.}
    \label{tab:avqa_general}
    \small
    \setlength{\tabcolsep}{3.5pt}
    \renewcommand{\arraystretch}{1.12}

    \begin{tabular}{@{\hspace{\tabcolsep}}lcc@{\hspace{\tabcolsep}}}
        \toprule

        \textbf{Method}
        & \textbf{OmniVideoBench}
        & \textbf{FutureOmni} \\

        \midrule

        VideoLLaMA2-AV
        & 27.5 & 38.8 \\

        VideoLLaMA2-AV+MAD
        & 28.3 & 36.3 \\

        \rowcolor{pacdpurple}[\tabcolsep][\tabcolsep]
        \textbf{VideoLLaMA2-AV+PACD}
        & \textbf{31.1\,{\footnotesize($\uparrow$9.9\%)}}
        & \textbf{40.0\,{\footnotesize($\uparrow$10.2\%)}} \\
        
        \midrule
        OmniVinci
        & 32.9 & 52.7 \\

        OmniVinci+MAD
        & 32.5 & 51.7 \\

        \rowcolor{pacdpurple}[\tabcolsep][\tabcolsep]
        \textbf{OmniVinci+PACD}
        & \textbf{34.8\,{\footnotesize($\uparrow$7.1\%)}}
        & \textbf{53.5\,{\footnotesize($\uparrow$3.5\%)}} \\

        \bottomrule
    \end{tabular}
\end{table}

\begin{table}[t]
    \centering
    \caption{Computational efficiency on CMM and AVHBench using VideoLLaMA2-AV-7B.}
    \label{tab:efficiency}
    \small
    \setlength{\tabcolsep}{3pt}
    \renewcommand{\arraystretch}{1.12}

    \begin{tabularx}{\columnwidth}{l*{4}{>{\centering\arraybackslash}X}}
        \toprule

        & \multicolumn{2}{c}{\textbf{CMM}}
        & \multicolumn{2}{c}{\textbf{AVHBench}} \\

        \cmidrule(lr){2-3}
        \cmidrule(lr){4-5}

        \textbf{Method}
        & \textbf{Latency}
        & \textbf{GPU Mem.}
        & \textbf{Latency}
        & \textbf{GPU Mem.} \\

        \midrule

        AVCD
        & 1.70s
        & 22.6G
        & 1.40s
        & 22.7G \\

        MAD
        & 1.41s
        & 20.7G
        & 1.40s
        & 20.7G \\

        \rowcolor{pacdpurple}
        \textbf{PACD}
        & 1.45s
        & 20.7G
        & 1.44s
        & 20.7G \\

        \bottomrule
    \end{tabularx}
\end{table}

\subsection{Generalization to Audio-Visual QA}
To evaluate PACD's generalization beyond hallucination-focused benchmarks, we further assess its performance on OmniVideoBench and FutureOmni using VideoLLaMA2-AV and OmniVinci. As shown in Table~\ref{tab:avqa_general}, PACD achieves the highest overall accuracy on both benchmarks across the two backbones. Specifically, PACD achieves relative improvements of 9.9\% and 10.2\% over MAD on VideoLLaMA2-AV, and 7.1\% and 3.5\% on OmniVinci, for OmniVideoBench and FutureOmni, respectively. These results demonstrate that PACD can also improve performance on broader audio-visual reasoning tasks.

\subsection{Computational Efficiency}
We evaluate the computational efficiency of PACD on CMM and AVHBench using VideoLLaMA2-AV-7B. As shown in Table~\ref{tab:efficiency}, PACD achieves inference latency comparable to MAD across both benchmarks while maintaining identical peak GPU memory usage. Notably, PACD consistently improves prediction accuracy over MAD with only marginal differences in inference time, demonstrating a favorable balance between effectiveness and efficiency. These results highlight the practical applicability of PACD for reliable audio-visual reasoning without substantially increasing computational cost.

\subsection{Robustness and Consensus Recovery}
\begin{figure}[t]
    \centering
    \includegraphics[width=1\linewidth]{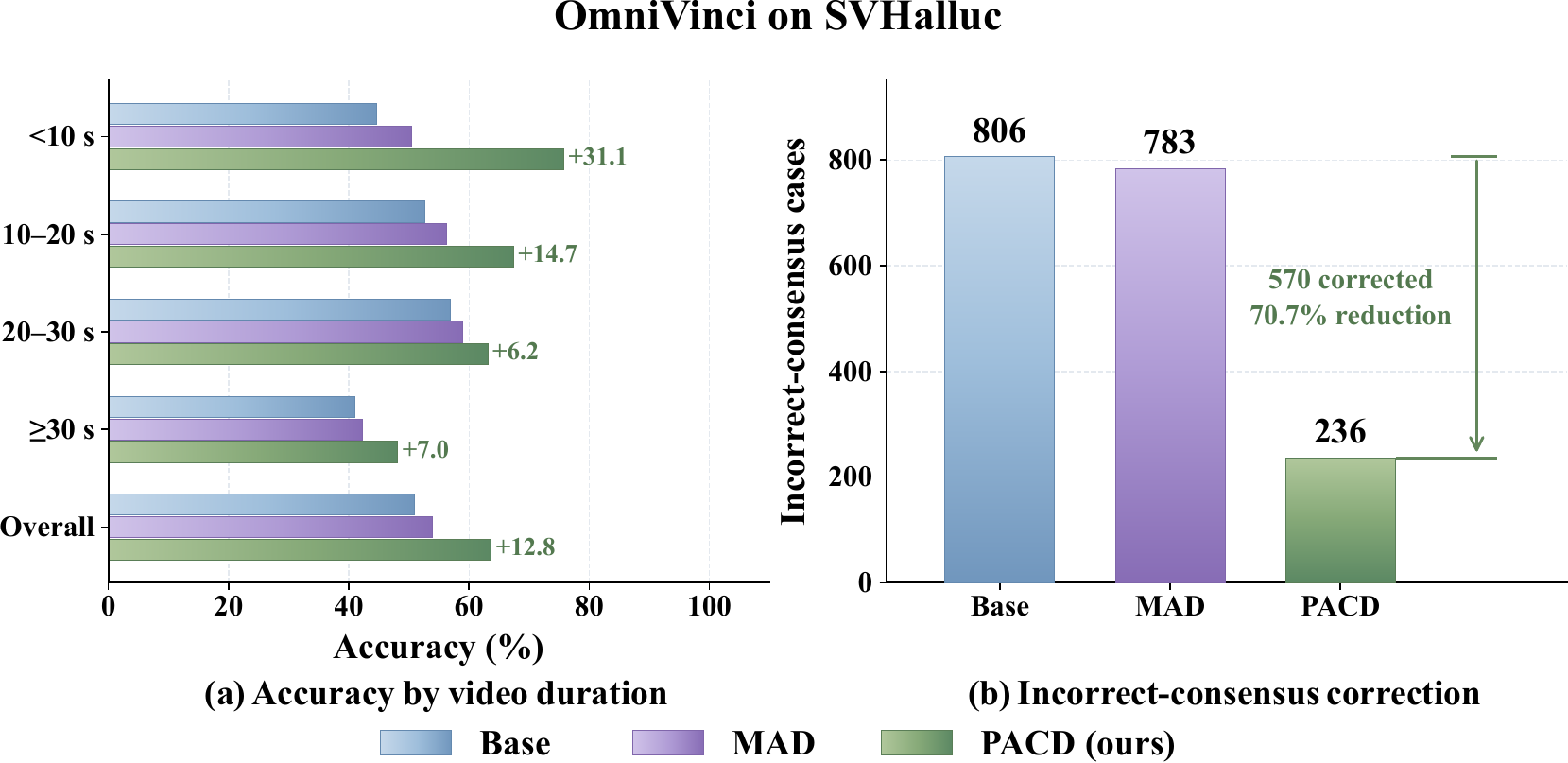}
    \caption{Performance across video durations and incorrect-consensus correction on SVHalluc using OmniVinci.}
    \label{fig:effectiveness}
\end{figure}

To evaluate PACD's robustness across different video durations, we conduct further analysis using OmniVinci on SVHalluc. As shown in Fig.~\ref{fig:effectiveness}(a), PACD consistently improves accuracy across all duration groups, achieving gains of 31.1, 14.7, 6.2, and 7.0 percentage points over Base, respectively. These results demonstrate its effectiveness on both short and long videos.
We further examine PACD's ability to correct incorrect consensus. As shown in Fig.~\ref{fig:effectiveness}(b), MAD corrects only 23 of the 806 incorrect-consensus cases, whereas PACD corrects 570, achieving a 70.7\% reduction. These results highlight PACD's effectiveness in resolving incorrect consensus and improving prediction reliability.

\section{Conclusion}
In this work, we identify two key challenges in existing methods for mitigating cross-modal hallucinations: the neglect of reliable modality-specific predictions and the persistence of erroneous consensus. To address these challenges, we propose Provenance-Aware Consensus Decoding (PACD), a training-free framework combining relevance-guided candidate arbitration with provenance-based consensus recovery. RGCA preserves task-relevant unimodal predictions through provenance-informed arbitration, while PBCR leverages the non-compensatory outranking principle to selectively resolve erroneous consensus using layer-wise evidence and confidence constraints. Experiments across four Omni-LLMs demonstrate consistent gains on CMM, AVHBench, and SVHalluc, alongside improved performance on OmniVideoBench and FutureOmni, highlighting PACD's effectiveness in hallucination mitigation and broader audio-visual reasoning.

\section*{LLM Usage Statement}
During the preparation of this manuscript, we used ChatGPT to assist with writing and language polishing across the entire manuscript, with the aim of improving the clarity and readability of the presentation. The authors have reviewed the AI-assisted text and take full responsibility for the final content of the paper, including the accuracy of its statements, the interpretation of the reported results, and the conclusions presented.
{
    \small
    \bibliographystyle{ieeenat_fullname}
    \bibliography{main}
}

\clearpage
\setcounter{page}{1}
\maketitlesupplementary

\section{Supplementary Evaluation Protocol}
\label{sec:supp_protocol}

This supplementary material provides additional evaluations of PACD, including comparisons with alternative decoding strategies, component ablations, fine-grained task analyses, hyperparameter sensitivity studies, duration-based evaluations, and statistical analyses. All experiments follow the model and method definitions established in the main paper, covering four representative Omni-LLMs: VideoLLaMA2-AV-7B, OmniVinci, Qwen2.5-Omni-3B, and Qwen2.5-Omni-7B. Evaluations are conducted on five benchmarks: CMM, AVHBench, SVHalluc, OmniVideoBench, and FutureOmni.

\paragraph{Evaluation units and aggregation.} CMM contains 800 questions, with 400 questions in each of the visual-dominant and audio-dominant categories. AVHBench contains 3,426 questions: 2,290 video-driven audio-hallucination questions and 1,136 audio-driven video-hallucination questions. SVHalluc contains 2,405 questions, comprising 1,422 semantic and 983 temporal questions. Its six tasks contain 444 GSA, 444 FGSA, 534 CMSB, 368 TA, 276 TF, and 339 CMTB questions. OmniVideoBench contains 1,000 multiple-choice questions from 628 videos, covering 13 reasoning types. FutureOmni contains 1,034 questions from 919 videos across eight domains, focusing on future-event forecasting through cross-modal causal and temporal reasoning. Unless stated otherwise, results are question-level accuracy in percent. Overall accuracy is weighted by the number of questions, rather than obtained by averaging the displayed category or task percentages. Table entries are rounded to one decimal place; boldface denotes the highest displayed value within each model block, including ties after rounding. Improvements are reported in percentage points.

\subsection{Backbone Models}
\label{sec:supp_backbones}

\paragraph{Model coverage.} We evaluate PACD using four pretrained Omni-LLMs that support joint audio-visual question answering: VideoLLaMA2-AV-7B, OmniVinci, Qwen2.5-Omni-3B, and Qwen2.5-Omni-7B. These backbones cover different model families and parameter scales, allowing us to examine whether provenance-guided decoding transfers across distinct audio-visual modeling architectures. All backbone parameters remain frozen throughout evaluation. PACD operates on their inference outputs and internal evidence without additional training or changes to the pretrained model weights.

\paragraph{VideoLLaMA2-AV-7B.} VideoLLaMA2-AV-7B is an audio-visual large language model built upon VideoLLaMA 2.1. It incorporates a Spatial-Temporal Convolution (STC) connector for video representation and an audio branch for joint audio-visual understanding. The model processes visual and auditory information alongside textual instructions to generate natural language responses. We use its audio-visual variant to evaluate PACD on tasks requiring modality-specific perception and cross-modal reasoning.

\paragraph{OmniVinci.} OmniVinci is a 9B-parameter omnimodal large language model designed for joint understanding of images, videos, audio, and text. Its architecture incorporates OmniAlignNet to align visual and auditory representations in a shared latent space, Temporal Embedding Grouping to capture relative temporal relationships, and Constrained Rotary Time Embedding to encode absolute temporal information. These designs support the integration of multimodal information for audio-visual perception and reasoning. We include OmniVinci to evaluate PACD on an architecture with explicit cross-modal alignment and temporal modeling mechanisms.

\paragraph{Qwen2.5-Omni-3B and Qwen2.5-Omni-7B.} Qwen2.5-Omni is an end-to-end omnimodal model capable of processing text, images, audio, and video. It adopts a Thinker--Talker architecture, where the Thinker performs multimodal perception, reasoning, and text generation, while the Talker supports speech generation. The model also introduces Time-aligned Multimodal Rotary Position Embedding (TMRoPE) to align temporal information across audio and video inputs. We evaluate PACD using the text prediction capabilities of the 3B and 7B variants, enabling comparisons across different model scales within the same architecture.

\subsection{Benchmark Descriptions}
\label{sec:supp_benchmarks}

\paragraph{CMM.} CMM examines hallucinations caused by misleading cross-modal cues and the dominance of one perceptual modality over another. Our evaluation focuses on its visual-dominant and audio-dominant categories. Visual-dominant questions assess predictions when visual evidence can interfere with judgments requiring audio information, while audio-dominant questions examine the corresponding influence of auditory evidence on visually grounded judgments. These categories provide a direct test of whether a decoding method can identify and preserve evidence from the modality relevant to the question. All evaluated CMM questions use binary answers.

\paragraph{AVHBench.} AVHBench evaluates hallucinations arising from interactions between audio and visual inputs. We use the video-driven audio-hallucination and audio-driven video-hallucination categories. The former examines judgments about audio in the presence of potentially misleading video information, whereas the latter examines judgments about video when auditory cues may affect the prediction. The two categories therefore evaluate complementary directions of cross-modal interference.

\paragraph{SVHalluc.} SVHalluc evaluates speech-vision understanding through three semantic and three temporal tasks. Global Semantic Alignment (GSA) concerns overall semantic correspondence between the speech and visual streams, while Fine-Grained Semantic Alignment (FGSA) examines more specific semantic relationships. Cross-Modal Semantic Binding (CMSB) focuses on associating information across modalities. The temporal tasks assess different aspects of temporal reasoning: Temporal Alignment (TA), Temporal Forecasting (TF), and Cross-Modal Temporal Binding (CMTB). This organization allows us to report performance separately for semantic correspondence, temporal relationships, and the overall benchmark. SVHalluc includes both two-choice and three-choice questions, which are handled using each question's valid candidate set.

\paragraph{OmniVideoBench.} OmniVideoBench evaluates general audio-visual understanding with an emphasis on modality complementarity and joint reasoning. It contains 1,000 question-answer pairs from 628 videos, covering 13 reasoning types, including temporal reasoning, spatial localization, counting, and causal inference. The questions require models to integrate information from both audio and visual streams rather than relying on a single modality. This benchmark allows us to examine whether PACD generalizes beyond hallucination-focused tasks to broader audio-visual reasoning.

\paragraph{FutureOmni.} FutureOmni evaluates future-event prediction from audio-visual context, requiring models to reason about causal and temporal relationships across modalities. It contains 1,034 multiple-choice questions from 919 videos across eight domains. Unlike conventional benchmarks that primarily assess understanding of observed events, FutureOmni focuses on predicting what may happen next based on available audio-visual evidence. We use this benchmark to evaluate PACD's generalization to predictive reasoning tasks.

\paragraph{Complementary evaluation settings.} The five benchmarks cover complementary aspects of audio-visual understanding. CMM focuses on modality dominance, AVHBench examines bidirectional cross-modal interference, and SVHalluc evaluates semantic and temporal hallucinations. OmniVideoBench extends the evaluation to general audio-visual reasoning, while FutureOmni further examines future-event prediction. Together, these benchmarks allow us to assess PACD's effectiveness in mitigating cross-modal hallucinations and its generalization to broader audio-visual reasoning tasks.

\subsection{Baseline Methods}
\label{sec:supp_baselines}

\paragraph{Base.} Base denotes the original joint audio-visual prediction produced by each pretrained backbone using its standard decoding procedure. The model receives the available audio, video, and textual question without any PACD intervention. We retain the original generated answer as the Base prediction, providing a reference for measuring improvements over standard audio-visual inference.

\paragraph{Audio-Visual Contrastive Decoding (AVCD).} AVCD is a training-free audio-visual hallucination mitigation method that constructs contrastive predictions using modality-dependent changes to the input evidence. By comparing the original audio-visual prediction with its contrastive counterparts, AVCD adjusts the decoding scores to reduce the influence of misleading modality information. We include AVCD as an established audio-visual contrastive decoding baseline in the main-paper comparisons.

\paragraph{Modality-Adaptive Decoding (MAD).} MAD is a training-free decoding method that dynamically adjusts the contributions of different modalities according to their relevance to the input query. It integrates predictions from joint audio-visual, audio-only, video-only, and text-only branches through a relevance-weighted contrastive decoding strategy to mitigate cross-modal hallucinations. We include MAD as a representative modality-adaptive decoding baseline and use a contrastive strength of $\gamma=2.5$, consistent with the main-paper experimental setting.

\paragraph{Additional decoding comparisons.} In addition to these established baselines, we evaluate Uniform and Argmax as simplified relevance-based strategies. Uniform removes question-dependent variation in the relevance weights, while Argmax converts the relevance scores into a one-hot distribution within the original MAD fusion rule. We also examine Routing, RGCA, and R+PBCR to separate the contributions of modality routing, candidate arbitration, and consensus recovery. Their exact prediction rules are provided in Secs.~\ref{sec:supp_relevance} and~\ref{sec:supp_components}. These comparisons distinguish PACD's full decision procedure from simpler uses of modality relevance and its individual components.

\subsection{Implementation and Prediction Protocol}
\label{sec:supp_implementation}

\paragraph{Inference setup.} All evaluations use frozen pretrained backbones, and PACD introduces no additional parameter training. The main-paper experiments are conducted on NVIDIA A100 GPUs with 80\,GB of memory. For CMM, AVHBench, and SVHalluc, video inputs are sampled at 2 frames per second with a maximum of 768 frames, following the main-paper setup. We use the same evaluated questions and input preparation for method comparisons within each backbone. The video-duration analyses additionally group questions by the durations of their prepared inputs, as described in Sec.~\ref{sec:supp_duration}.

\paragraph{Answer candidates.} Each question $i$ is associated with a valid answer set $\mathcal{C}_i$. For CMM and AVHBench, the candidate set consists of the binary answers \textit{Yes} and \textit{No}; SVHalluc uses the two or three choices defined for each question. For each question $i$, $\phi_i(c)$ denotes the answer-token index corresponding to candidate $c$ under the backbone tokenizer. Candidate scores are obtained from the logits of these mapped tokens at the answer-generation step. The same valid candidate set $\mathcal{C}_i$ is used for comparisons across all modality-conditioned branches and MAD, as well as for the final PACD answer selection.

\paragraph{Modality-conditioned predictions.} We distinguish the joint audio-visual ($AV$), audio-only ($A$), video-only ($V$), and text-only ($T$) input conditions. Each branch retains the same textual question while changing which perceptual inputs are available. The predictions $c_A$, $c_V$, and $c_{AV}$ provide the candidate-level information used for routing and consensus assessment, while the associated logits support confidence comparisons. PACD combines these outputs with provenance information extracted from the joint audio-visual computation, following the definitions in the main paper.

\section{Additional Method Details}
\label{sec:supp_method_details}

This section provides further explanation of the candidate selection and decision process underlying PACD. The method consists of Relevance Guided Candidate Arbitration (RGCA) and Provenance Based Consensus Recovery (PBCR). Both components use the same frozen backbone and the modality-conditioned predictions defined in the main paper. RGCA focuses on retaining reliable modality-specific candidates, while PBCR uses layer-wise evidence to selectively recover a supported alternative when the audio-only, video-only, and joint audio-visual branches agree.

\subsection{Relevance-Guided Routing}
\label{sec:supp_routing_details}

PACD begins with the modality relevance scores $r_A$, $r_V$, and $r_{AV}$ obtained from the MAD relevance assessment. RGCA compares the two unimodal scores to identify the target modality: audio is selected when $r_A>r_V$, and video is selected otherwise. The corresponding unimodal prediction becomes the routed candidate $c_R$. Importantly, $c_R$ is the original predicted answer from the selected branch, rather than a new answer obtained by ranking the candidates with a different scoring rule.

This distinction separates the identification of relevant evidence from the decision to adopt that evidence as the final answer. The relevance scores indicate which perceptual stream is most closely associated with the question, while the subsequent candidate-level comparison considers the predictive support for the routed answer. RGCA therefore uses relevance to guide candidate selection and confidence-based arbitration to determine how the routed candidate should interact with the original MAD prediction.

\subsection{Layer-Wise Provenance Evidence}
\label{sec:supp_provenance_details}

After identifying $c_R$, PACD considers the remaining candidates in $\mathcal{C}_i\setminus\{c_R\}$. To evaluate these alternatives, it examines modality-specific attention contributions from the existing joint audio-visual forward computation. Audio and video token positions are identified within the joint input, and their attention contributions are projected into the candidate-logit space using the backbone's output projection. This provides a candidate-level signal of how the selected modality contributes to the prediction across decoder layers.

For an alternative candidate $c$, the provenance difference at layer $\ell$ measures the relative contribution to $c$ and $c_R$ from the target modality. The differences are normalized using a robust contribution scale and aggregated over the last $N$ recorded layers. Their median, denoted by $Q(c)$, summarizes the strength of the layer-wise support for each alternative. PACD selects the candidate $c^*$ with the highest median provenance score among the valid alternatives.

PBCR also uses the provenance consistency score $C(c^*)$, which measures the fraction of examined layers in which the target modality favors $c^*$ over $c_R$. The statistics $Q(c^*)$ and $C(c^*)$ capture complementary properties of the same evidence: the magnitude of the alternative's support and its consistency across layers. Both are computed from the recorded provenance information, allowing PACD to assess candidate evidence without modifying the backbone's attention outputs or hidden states.

\subsection{Confidence-Based Candidate Arbitration}
\label{sec:supp_arbitration_details}

RGCA compares the routed candidate $c_R$ with the provenance-supported alternative $c^*$ using two normalized confidence margins. The target-branch margin $T$ measures the preference of the selected unimodal branch for $c_R$ over $c^*$, while the MAD margin $M$ measures the corresponding preference under the original MAD decoding scores. Both margins are normalized by a robust scale derived from candidate-logit differences across the available branches, making their comparisons less sensitive to backbone-dependent score magnitudes.

The confidence veto in RGCA combines these margins with the relationship between $c_R$ and $c_{\mathrm{MAD}}$. When the veto condition holds, the arbitration output $c_B$ retains the MAD prediction; otherwise, it adopts the routed candidate. This step brings together the question's modality relevance, the selected branch's own prediction, and the confidence expressed by MAD. The result $c_B$ provides the candidate selected by RGCA before any consensus recovery is considered.

\subsection{Provenance-Based Consensus Recovery}
\label{sec:supp_recovery_details}

PBCR operates when the audio-only, video-only, and joint audio-visual branches reach consensus, identified by $c_A=c_V=c_{AV}$. Since branch agreement may still conceal shared errors, PBCR examines whether internal provenance evidence supports an alternative candidate $c^*$. The recovery decision combines candidate uniqueness, cross-layer provenance consistency, provenance strength, and predictive confidence to determine whether the shared prediction should be corrected.

The uniqueness condition $U_i$ ensures that $c^*$ is distinguishable from other valid alternatives. The provenance consistency score $C(c^*)$ and median support $Q(c^*)$ are evaluated against thresholds $\tau_C$ and $\tau_S$, respectively. Meanwhile, the normalized confidence margins $T$ and $M$ constrain recovery through $\max(T,M)<\tau_R$. As defined in Eq.~\eqref{eq:pacd_rescue_gate}, recovery is triggered only when all these conditions are satisfied.

The final PACD prediction follows Eq.~\eqref{eq:pacd_final_prediction}. For non-consensus cases, PACD adopts the RGCA arbitration result $c_B$. For consensus cases, PBCR selects $c^*$ only when the recovery conditions are satisfied; otherwise, PACD retains the original MAD prediction $c_{\mathrm{MAD}}$. This conditional decision mechanism allows PACD to preserve reliable modality-specific predictions while selectively correcting incorrect consensus.

\subsection{Complementary Roles of RGCA and PBCR}
\label{sec:supp_complementarity}

RGCA and PBCR address two related aspects of cross-modal hallucination. RGCA focuses on cases where a task-relevant unimodal branch provides a useful answer that should be considered alongside the fused prediction. PBCR focuses on cases where agreement among the modality-conditioned branches can be revisited using alternative candidate evidence. The two components share the routed candidate and provenance statistics, but apply them to different decisions: arbitration determines which current prediction to retain, whereas recovery assesses whether a supported alternative should take precedence.

This separation is reflected in the component experiments in Sec.~\ref{sec:supp_components}. By combining relevance-guided routing, candidate confidence, and layer-wise provenance, PACD uses both modality-specific predictions and internal audio-visual evidence within a unified training-free decoding procedure.

\section{Hyperparameter Sensitivity}
\label{sec:supp_sensitivity}

Table~\ref{tab:pacd_controls} summarizes the PACD parameters and numerical controls. The listed hyperparameter values define a reference configuration. We select hyperparameters using approximately 10\% of the samples from each benchmark through coarse-to-fine search, without updating model weights.

\begin{table}[htbp]
\centering
\caption{PACD hyperparameters and their implementation roles, with the reference configuration $N=10$, $\tau_T=1.2$, $\tau_M=3.0$, $\tau_C=0.2$, $\tau_S=-0.6$, and $\tau_R=2.8$.}
\label{tab:pacd_controls}
\small
\begin{tabular}{@{}p{0.20\linewidth}p{0.74\linewidth}@{}}
\toprule
Control & Meaning and implementation role \\
\midrule

$N$
& Number of decoder layers included in provenance evidence aggregation. \\

$\tau_T$
& Confidence threshold for determining the preference of the routed target branch in RGCA. \\

$\tau_M$
& Confidence threshold for determining the preference of the MAD prediction in RGCA. \\

$\tau_C$
& Minimum cross-layer provenance consistency required for PBCR recovery. \\

$\tau_S$
& Minimum provenance support strength required for PBCR recovery. \\

$\tau_R$
& Confidence threshold governing selective consensus recovery in PBCR. \\

\bottomrule
\end{tabular}
\end{table}

\subsection{Number of Examined Layers}
\label{sec:supp_layers}

\begin{figure}[t]
    \centering
    \includegraphics[width=\columnwidth]{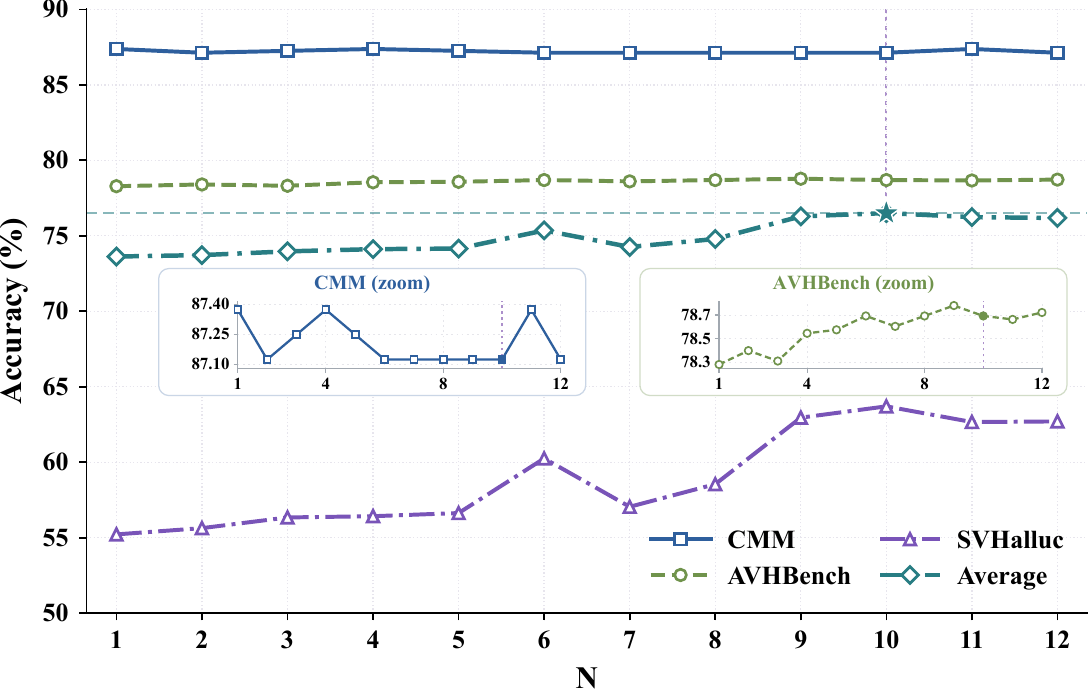}
    \caption{Sensitivity to the number of examined layers $N$ on CMM, AVHBench, and SVHalluc using OmniVinci. Insets provide enlarged views of the accuracy variations on CMM and AVHBench.}
    \label{fig:ablation_N}
\end{figure}

Figure~\ref{fig:ablation_N} investigates the effect of the number of examined layers $N$ on PACD. We vary $N$ from 1 to 12 while keeping all other hyperparameters fixed. Specifically, $\mathcal{L}_N$ comprises the last $N$ layers within the available 12-layer recording window.

As shown in Fig.~\ref{fig:ablation_N}, PACD maintains stable accuracy on CMM and AVHBench across different values of $N$. On SVHalluc, aggregating evidence across more layers generally leads to improved accuracy, with the best performance achieved at $N=10$. Notably, the average accuracy across the three benchmarks also reaches its maximum at $N=10$, demonstrating that this configuration provides an effective aggregation window for capturing complementary layer-wise provenance evidence. The reference configuration therefore uses $N=10$.

\subsection{RGCA Threshold Sensitivity}
\label{sec:supp_rgca_thresholds}

\begin{figure}[t]
    \centering
    \includegraphics[width=\columnwidth]{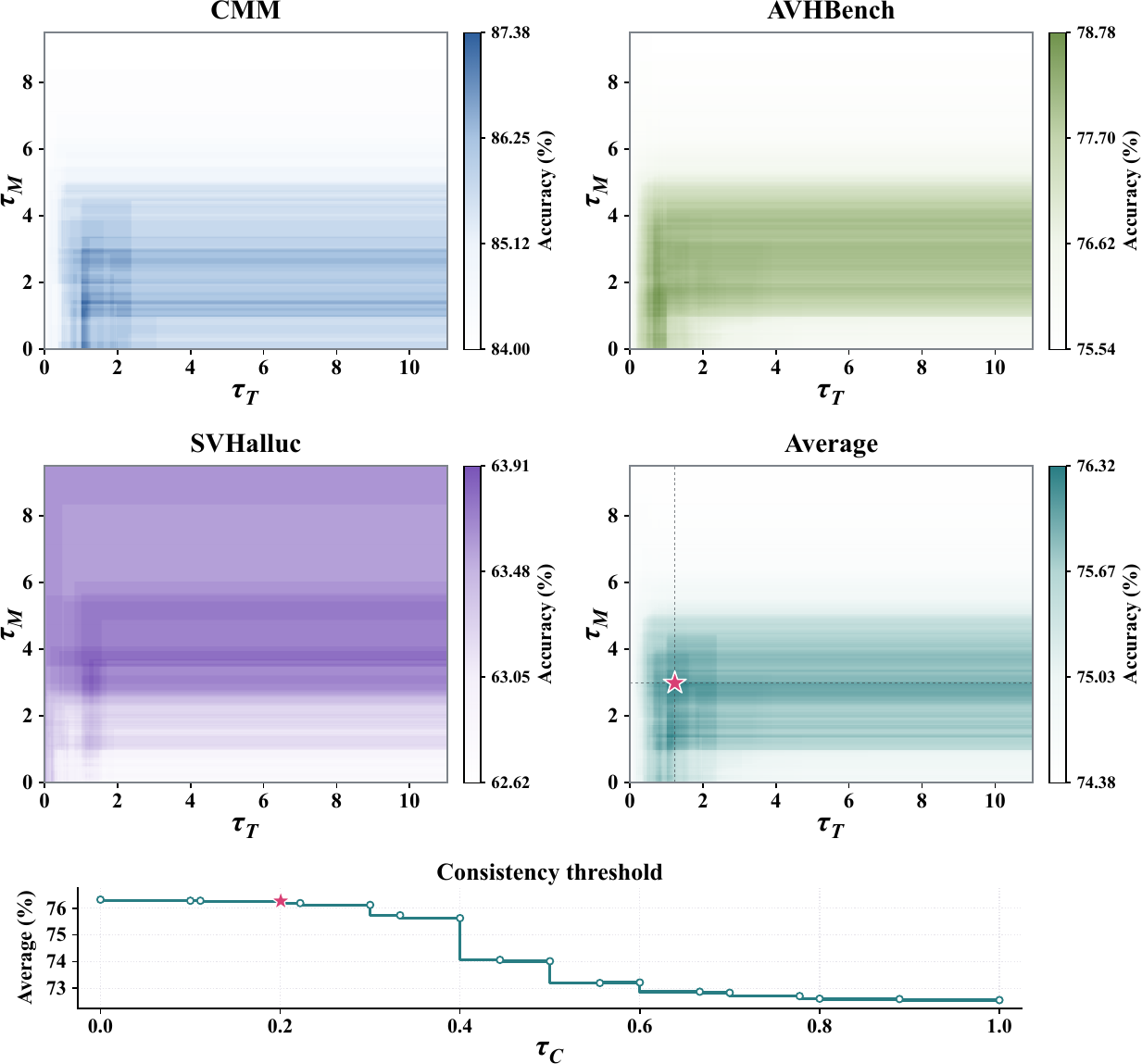}
    \caption{Hyperparameter sensitivity of RGCA and PBCR on CMM, AVHBench, and SVHalluc using OmniVinci. The upper heatmaps show the joint effects of $\tau_T$ and $\tau_M$, while the bottom plot shows the average accuracy under different $\tau_C$ values. Stars indicate the reference configurations.}
    \label{fig:ablation_tauT_tauM_tauC}
\end{figure}

Figure~\ref{fig:ablation_tauT_tauM_tauC} (top) examines the joint effects of the two confidence thresholds in RGCA, $\tau_T$ and $\tau_M$. We vary both thresholds while keeping the remaining hyperparameters fixed. The results reveal distinct accuracy distributions across the three benchmarks, with broad high-performing regions. Specifically, $\tau_T$ primarily governs the target-branch preference, while $\tau_M$ regulates the preference of the MAD prediction, allowing RGCA to adapt its candidate arbitration to different benchmark characteristics.

The average accuracy map identifies a high-performing region around $\tau_T=1.2$ and $\tau_M=3.0$, corresponding to the reference configuration. These values provide a favorable balance across the three benchmarks.

\subsection{Provenance Strength and Recovery Confidence}
\label{sec:supp_pbcr_thresholds}

\begin{figure}[t]
    \centering
    \includegraphics[width=\columnwidth]{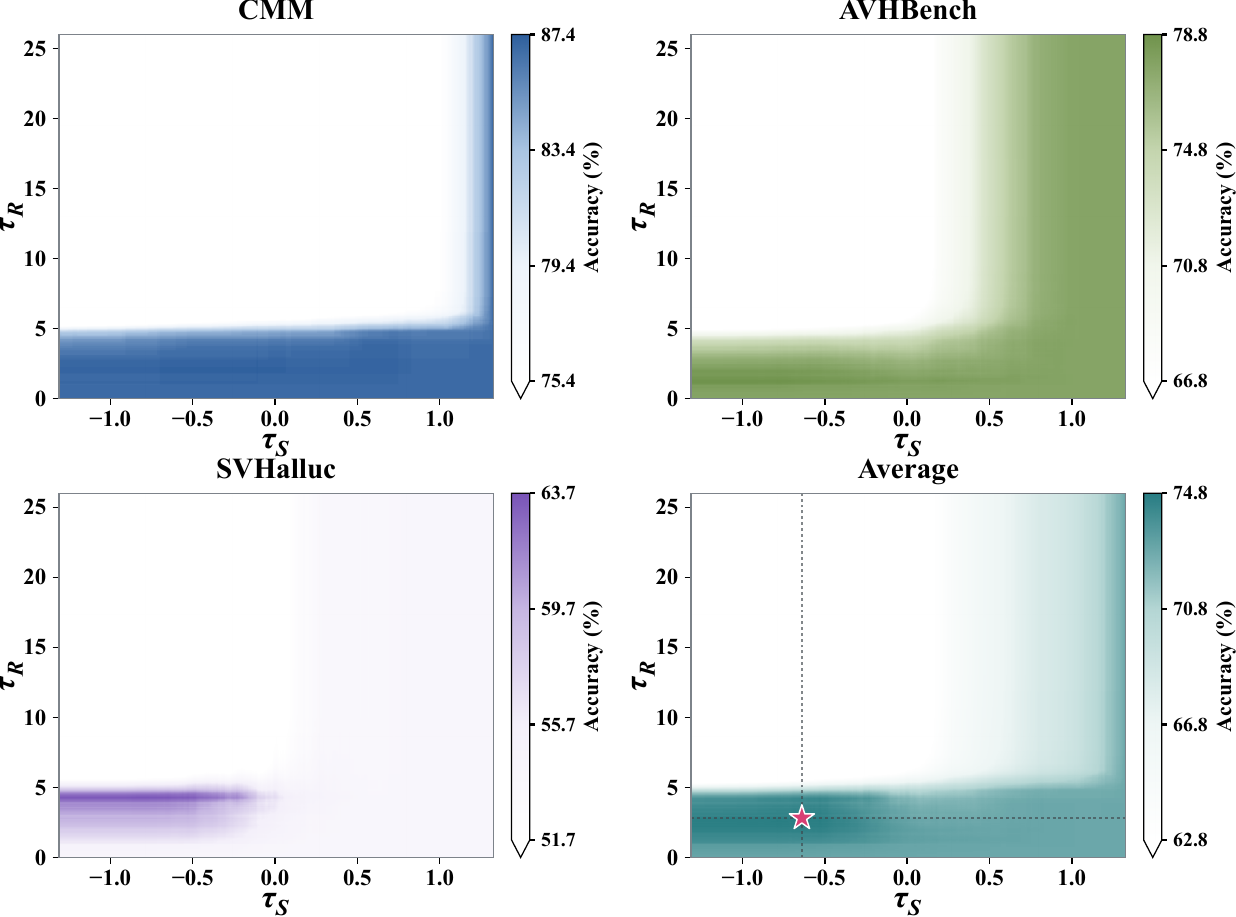}
    \caption{Joint sensitivity of PBCR to the provenance-strength threshold $\tau_S$ and recovery-confidence threshold $\tau_R$ on CMM, AVHBench, and SVHalluc using OmniVinci. The star indicates the reference configuration $\tau_S=-0.6$ and $\tau_R=2.8$.}
    \label{fig:ablation_tauS_tauR}
\end{figure}

Figure~\ref{fig:ablation_tauS_tauR} examines the joint effects of the provenance-strength threshold $\tau_S$ and recovery-confidence threshold $\tau_R$ in PBCR. We vary these two thresholds simultaneously while keeping the remaining hyperparameters fixed. The heatmaps illustrate how provenance support and prediction confidence jointly determine the recovery behavior across different benchmarks. In particular, the high-performing regions demonstrate that PBCR can effectively identify suitable alternative candidates by combining provenance strength with confidence-based constraints.

The average accuracy map identifies a favorable joint configuration at $\tau_S=-0.6$ and $\tau_R=2.8$, corresponding to the reference configuration. These values support selective consensus recovery based on complementary provenance and confidence evidence.

\begin{table*}[t]
    \centering
    \caption{\textbf{Complete comparison with simplified decoding strategies.} Accuracy (\%) on CMM, AVHBench, and SVHalluc for all four backbones. Uniform assigns equal modality relevance within the MAD combination rule; Argmax converts modality relevance into a one-hot assignment and applies the same MAD combination rule. Base denotes the original joint audio-visual generation output. Overall accuracy is question-weighted. Bold indicates the highest displayed result within each model and metric, including ties after rounding.}

    \label{tab:fusion_ablation_full}

    \small
    \setlength{\tabcolsep}{0.5mm}
    \renewcommand{\arraystretch}{0.98}

    \begin{tabular*}{\textwidth}{
        @{\extracolsep{\fill}}
        p{1.30cm}
        C{1.70cm}
        C{1.70cm}
        C{1.70cm}
        C{2.05cm}
        C{2.05cm}
        C{2.05cm}
        C{1.05cm}
        C{1.05cm}
        C{2.10cm}
        @{}
    }
        \toprule

        \multirow[c]{2}{*}{\raisebox{-0.98em}[0pt][0pt]{\textbf{Method}}}
        & \multicolumn{3}{c}{\textbf{CMM}}
        & \multicolumn{3}{c}{\textbf{AVHBench}}
        & \multicolumn{3}{c}{\textbf{SVHalluc}} \\

        \cmidrule(lr){2-4}
        \cmidrule(lr){5-7}
        \cmidrule(lr){8-10}

        &
        \parbox[c][2.6em][c]{1.70cm}{
            \centering\textbf{Visual Dom.}
        }
        &
        \parbox[c][2.6em][c]{1.70cm}{
            \centering\textbf{Audio Dom.}
        }
        &
        \parbox[c][2.6em][c]{1.70cm}{
            \centering\textbf{Overall Acc.}
        }
        &
        \parbox[c][2.6em][c]{2.05cm}{
            \centering
            \textbf{Video-Driven}\\
            \textbf{Audio Hall.}
        }
        &
        \parbox[c][2.6em][c]{2.05cm}{
            \centering
            \textbf{Audio-Driven}\\
            \textbf{Video Hall.}
        }
        &
        \parbox[c][2.6em][c]{2.05cm}{
            \centering\textbf{Overall Acc.}
        }
        &
        \parbox[c][2.6em][c]{1.05cm}{
            \centering\textbf{Sem.}
        }
        &
        \parbox[c][2.6em][c]{1.05cm}{
            \centering\textbf{Temp.}
        }
        &
        \parbox[c][2.6em][c]{2.10cm}{
            \centering\mbox{\textbf{Overall Acc.}}
        }
        \\

        \midrule

        \multicolumn{10}{c}{\textbf{VideoLLaMA2-AV-7B}} \\
        \midrule

        Base
        & 72.3 & 82.3 & 77.3
        & 70.8 & 76.1 & 72.6
        & 57.2 & 43.8 & 51.7 \\

        Uniform
        & 77.5 & \textbf{82.8} & 80.1
        & 77.1 & 77.2 & 77.1
        & 61.5 & 43.9 & 54.3 \\

        Argmax
        & 80.0 & 71.3 & 75.6
        & 77.4 & 75.9 & 76.9
        & 64.1 & 44.5 & 56.0 \\

        MAD
        & 81.3 & 80.5 & 80.9
        & 79.7 & 77.0 & 78.8
        & 65.1 & 44.5 & 56.7 \\

        \rowcolor{pacdpurple}
        \textbf{PACD}
        & \textbf{86.5} & 78.3 & \textbf{82.4}
        & \textbf{81.2} & \textbf{79.4} & \textbf{80.6}
        & \textbf{71.0} & \textbf{45.0} & \textbf{60.4} \\

        \midrule
        \multicolumn{10}{c}{\textbf{OmniVinci}} \\
        \midrule

        Base
        & 81.0 & 83.8 & 82.4
        & 72.2 & 68.8 & 71.1
        & 54.5 & 45.8 & 50.9 \\

        Uniform
        & 86.0 & 85.8 & 85.9
        & 76.2 & 68.6 & 73.7
        & 57.0 & 46.2 & 52.6 \\

        Argmax
        & 86.3 & 82.3 & 84.3
        & 74.7 & 68.8 & 72.7
        & 59.6 & 45.1 & 53.6 \\

        MAD
        & 86.0 & 84.8 & 85.4
        & 77.6 & 70.2 & 75.1
        & 59.2 & 46.3 & 53.9 \\

        \rowcolor{pacdpurple}
        \textbf{PACD}
        & \textbf{88.5} & \textbf{86.3} & \textbf{87.4}
        & \textbf{82.6} & \textbf{71.0} & \textbf{78.8}
        & \textbf{75.7} & \textbf{46.4} & \textbf{63.7} \\

        \midrule
        \multicolumn{10}{c}{\textbf{Qwen2.5-Omni-3B}} \\
        \midrule

        Base
        & 70.8 & 85.3 & 78.0
        & 73.4 & 80.5 & 75.8
        & 72.7 & 42.1 & 60.2 \\

        Uniform
        & 72.5 & 86.8 & 79.6
        & 75.9 & \textbf{82.7} & 78.2
        & 75.5 & 43.4 & 62.4 \\

        Argmax
        & \textbf{81.3} & 85.3 & 83.3
        & 81.2 & 80.2 & 80.9
        & 75.5 & 44.6 & 62.8 \\

        MAD
        & 74.5 & 86.3 & 80.4
        & 80.1 & 82.5 & 80.9
        & 77.9 & 45.2 & 64.5 \\

        \rowcolor{pacdpurple}
        \textbf{PACD}
        & 80.8 & \textbf{88.5} & \textbf{84.6}
        & \textbf{82.7} & 81.8 & \textbf{82.4}
        & \textbf{78.0} & \textbf{47.6} & \textbf{65.6} \\

        \midrule
        \multicolumn{10}{c}{\textbf{Qwen2.5-Omni-7B}} \\
        \midrule

        Base
        & 66.5 & 70.3 & 68.4
        & 73.0 & 80.4 & 75.5
        & 77.5 & 46.2 & 64.7 \\

        Uniform
        & 69.8 & 73.3 & 71.5
        & 74.5 & 83.9 & 77.6
        & 78.7 & 45.9 & 65.3 \\

        Argmax
        & 83.0 & 89.0 & 86.0
        & 78.8 & 84.3 & 80.6
        & \textbf{83.3} & 45.7 & 67.9 \\

        MAD
        & 76.5 & 84.0 & 80.3
        & 78.9 & \textbf{84.7} & 80.8
        & 82.3 & 46.7 & 67.8 \\

        \rowcolor{pacdpurple}
        \textbf{PACD}
        & \textbf{85.0} & \textbf{89.5} & \textbf{87.3}
        & \textbf{80.7} & \textbf{84.7} & \textbf{82.0}
        & 82.8 & \textbf{48.5} & \textbf{68.8} \\

        \bottomrule
    \end{tabular*}
\end{table*}

\begin{table*}[t]
    \centering
    \caption{\textbf{SVHalluc task-level comparison with AVCD and MAD.} Accuracy (\%) on all six tasks. GSA, FGSA, and CMSB denote Global Semantic Alignment, Fine-Grained Semantic Alignment, and Cross-Modal Semantic Binding; TA, TF, and CMTB denote Temporal Alignment, Temporal Forecasting, and Cross-Modal Temporal Binding. The first three tasks form the semantic category (Sem.) and the last three form the temporal category (Temp.). Avg. is question-weighted accuracy over all 2,405 questions. Bold denotes the highest displayed result within each backbone and column, including ties after rounding.}
    \label{tab:svhalluc_detailed}
    \small
    \setlength{\tabcolsep}{3pt}
    \renewcommand{\arraystretch}{1.02}
    \begin{tabularx}{\textwidth}{@{}>{\raggedright\arraybackslash}p{1.5cm}*{7}{>{\centering\arraybackslash}X}@{}}
        \toprule
        \multirow{2}{*}{\textbf{Method}} & \multicolumn{3}{c}{\textbf{Sem.}} & \multicolumn{3}{c}{\textbf{Temp.}} & \multirow{2}{*}{\textbf{Avg.}} \\
        \cmidrule(lr){2-4}\cmidrule(lr){5-7}
        & \textbf{GSA} & \textbf{FGSA} & \textbf{CMSB} & \textbf{TA} & \textbf{TF} & \textbf{CMTB} & \\
        \midrule
        \multicolumn{8}{c}{\textbf{VideoLLaMA2-AV-7B}} \\
        \midrule
        Base & \textbf{50.0} & 57.2 & 63.1 & \textbf{50.0} & 26.8 & 51.0 & 51.7 \\
        AVCD & \textbf{50.0} & 67.1 & 69.1 & \textbf{50.0} & 29.7 & \textbf{52.5} & 55.4 \\
        MAD & \textbf{50.0} & 69.6 & 74.0 & \textbf{50.0} & 29.7 & 50.4 & 56.7 \\
        \rowcolor{pacdpurple}
        \textbf{PACD} & \textbf{50.0} & \textbf{82.4} & \textbf{79.0} & \textbf{50.0} & \textbf{32.6} & 49.6 & \textbf{60.4} \\
        \midrule
        \multicolumn{8}{c}{\textbf{OmniVinci}} \\
        \midrule
        Base & 50.0 & 51.1 & 61.0 & 50.0 & \textbf{34.1} & 50.7 & 50.9 \\
        AVCD & 49.8 & 71.2 & \textbf{80.5} & 48.1 & 32.2 & 51.6 & 58.5 \\
        MAD & 50.0 & 55.4 & 70.0 & 50.0 & 33.0 & \textbf{53.1} & 53.9 \\
        \rowcolor{pacdpurple}
        \textbf{PACD} & \textbf{68.7} & \textbf{79.5} & 78.3 & \textbf{51.9} & 33.3 & 51.0 & \textbf{63.7} \\
        \midrule
        \multicolumn{8}{c}{\textbf{Qwen2.5-Omni-3B}} \\
        \midrule
        Base & \textbf{72.1} & 72.3 & 73.6 & 49.2 & 22.8 & 50.1 & 60.2 \\
        AVCD & 71.2 & 57.9 & 69.1 & \textbf{52.7} & 27.2 & \textbf{51.9} & 57.7 \\
        MAD & 68.2 & 81.8 & 82.8 & 50.0 & 32.2 & 50.4 & 64.5 \\
        \rowcolor{pacdpurple}
        \textbf{PACD} & 64.6 & \textbf{84.2} & \textbf{83.9} & 51.4 & \textbf{39.9} & 49.9 & \textbf{65.6} \\
        \midrule
        \multicolumn{8}{c}{\textbf{Qwen2.5-Omni-7B}} \\
        \midrule
        Base & 72.5 & 84.0 & 76.2 & 52.7 & 33.3 & 49.6 & 64.7 \\
        AVCD & \textbf{78.8} & 79.1 & 71.5 & 49.2 & 33.0 & \textbf{51.0} & 63.5 \\
        MAD & 71.4 & 89.2 & 85.8 & 53.3 & 33.3 & 50.4 & 67.8 \\
        \rowcolor{pacdpurple}
        \textbf{PACD} & 70.7 & \textbf{89.9} & \textbf{86.9} & \textbf{53.8} & \textbf{38.4} & \textbf{51.0} & \textbf{68.8} \\
        \bottomrule
    \end{tabularx}
\end{table*}

\begin{table*}[t]
    \centering
    \caption{\textbf{Task-level comparison of simplified decoding strategies on SVHalluc.} Uniform assigns equal modality relevance, while Argmax converts relevance into a one-hot assignment; both apply the MAD combination rule. Task abbreviations follow Table~\ref{tab:svhalluc_detailed}; Avg. is accuracy over all 2,405 questions, rather than the unweighted mean of the six displayed task accuracies. All values are percentages. Bold denotes the highest displayed result within each backbone and column, including ties after rounding.}
    \label{tab:svhalluc_detailed_fusion_ablation}
    \small
    \setlength{\tabcolsep}{3pt}
    \renewcommand{\arraystretch}{1.02}
    \begin{tabularx}{\textwidth}{@{}>{\raggedright\arraybackslash}p{1.5cm}*{7}{>{\centering\arraybackslash}X}@{}}
        \toprule
        \multirow{2}{*}{\textbf{Method}} & \multicolumn{3}{c}{\textbf{Sem.}} & \multicolumn{3}{c}{\textbf{Temp.}} & \multirow{2}{*}{\textbf{Avg.}} \\
        \cmidrule(lr){2-4}\cmidrule(lr){5-7}
        & \textbf{GSA} & \textbf{FGSA} & \textbf{CMSB} & \textbf{TA} & \textbf{TF} & \textbf{CMTB} & \\
        \midrule
        \multicolumn{8}{c}{\textbf{VideoLLaMA2-AV-7B}} \\
        \midrule
        Base & \textbf{50.0} & 57.2 & 63.1 & \textbf{50.0} & 26.8 & \textbf{51.0} & 51.7 \\
        Uniform & \textbf{50.0} & 65.1 & 68.0 & \textbf{50.0} & 29.3 & 49.3 & 54.3 \\
        Argmax & \textbf{50.0} & 60.1 & \textbf{79.0} & \textbf{50.0} & 29.0 & \textbf{51.0} & 56.0 \\
        MAD & \textbf{50.0} & 69.6 & 74.0 & \textbf{50.0} & 29.7 & 50.4 & 56.7 \\
        \rowcolor{pacdpurple}
        \textbf{PACD} & \textbf{50.0} & \textbf{82.4} & \textbf{79.0} & \textbf{50.0} & \textbf{32.6} & 49.6 & \textbf{60.4} \\
        \midrule
        \multicolumn{8}{c}{\textbf{OmniVinci}} \\
        \midrule
        Base & 50.0 & 51.1 & 61.0 & 50.0 & \textbf{34.1} & 50.7 & 50.9 \\
        Uniform & 50.0 & 53.4 & 65.9 & 50.0 & 32.2 & \textbf{53.4} & 52.6 \\
        Argmax & 50.0 & 51.6 & 74.3 & 50.0 & 33.3 & 48.7 & 53.6 \\
        MAD & 50.0 & 55.4 & 70.0 & 50.0 & 33.0 & 53.1 & 53.9 \\
        \rowcolor{pacdpurple}
        \textbf{PACD} & \textbf{68.7} & \textbf{79.5} & \textbf{78.3} & \textbf{51.9} & 33.3 & 51.0 & \textbf{63.7} \\
        \midrule
        \multicolumn{8}{c}{\textbf{Qwen2.5-Omni-3B}} \\
        \midrule
        Base & \textbf{72.1} & 72.3 & 73.6 & 49.2 & 22.8 & 50.1 & 60.2 \\
        Uniform & 65.3 & 79.5 & 80.7 & 50.8 & 27.9 & 48.1 & 62.4 \\
        Argmax & 59.9 & 82.0 & 83.0 & 48.6 & 33.3 & 49.3 & 62.8 \\
        MAD & 68.2 & 81.8 & 82.8 & 50.0 & 32.2 & \textbf{50.4} & 64.5 \\
        \rowcolor{pacdpurple}
        \textbf{PACD} & 64.6 & \textbf{84.2} & \textbf{83.9} & \textbf{51.4} & \textbf{39.9} & 49.9 & \textbf{65.6} \\
        \midrule
        \multicolumn{8}{c}{\textbf{Qwen2.5-Omni-7B}} \\
        \midrule
        Base & \textbf{72.5} & 84.0 & 76.2 & 52.7 & 33.3 & 49.6 & 64.7 \\
        Uniform & 71.4 & 86.0 & 78.7 & 50.5 & 33.3 & \textbf{51.0} & 65.3 \\
        Argmax & \textbf{72.5} & 89.4 & \textbf{87.1} & 52.7 & 33.3 & 48.1 & 67.9 \\
        MAD & 71.4 & 89.2 & 85.8 & 53.3 & 33.3 & 50.4 & 67.8 \\
        \rowcolor{pacdpurple}
        \textbf{PACD} & 70.7 & \textbf{89.9} & 86.9 & \textbf{53.8} & \textbf{38.4} & \textbf{51.0} & \textbf{68.8} \\
        \bottomrule
    \end{tabularx}
\end{table*}

\subsection{Cross-Layer Consistency Threshold}
\label{sec:supp_consistency}

We further investigate the effect of the cross-layer provenance consistency threshold $\tau_C$, which determines the minimum consistency required for an alternative candidate to be considered for consensus recovery. As shown in the bottom plot of Fig.~\ref{fig:ablation_tauT_tauM_tauC}, the average accuracy reaches its highest value at $\tau_C=0.2$, with a broad high-performing region around this value. This result demonstrates the effectiveness of incorporating cross-layer consistency into the consensus recovery process, ensuring that alternative candidates are supported by consistent provenance evidence. Accordingly, $\tau_C=0.2$ is adopted as the reference configuration.

\section{Additional Comparisons with Decoding Strategies}
\label{sec:supp_decoding}

\subsection{Modality Relevance and Branch Selection}
\label{sec:supp_relevance}

Table~\ref{tab:fusion_ablation_full} extends the relevance-weighting comparison to all four backbones and all benchmark categories. Uniform assigns equal relevance to all modalities, setting $r_A=r_V=r_{AV}=1/3$. Both Uniform and Argmax apply the resulting MAD decoding weights with the same contrastive strength $\gamma$, without candidate arbitration or consensus recovery.

Argmax converts the original relevance distribution into a one-hot assignment. Specifically, the selected modality and modified relevance scores are defined as
\begin{equation}
\begin{aligned}
b^* &= \operatorname*{arg\,max}_{b\in\{A,V,AV\}} r_b,\\
\tilde{r}_b &= \mathbb{I}[b=b^*], \quad b\in\{A,V,AV\}.
\end{aligned}
\label{eq:supp_argmax}
\end{equation}
The modified relevance scores $\tilde{r}_b$ are then substituted into the original MAD weighting rule in Eq.~\eqref{eq:mad_compact_decoding}. Thus, Argmax retains the MAD fusion structure rather than directly selecting a single branch prediction.

Table~\ref{tab:fusion_ablation_full} shows that PACD achieves the highest overall accuracy across all 12 model--benchmark combinations. On OmniVinci, PACD reaches 87.4\%, 78.8\%, and 63.7\% on CMM, AVHBench, and SVHalluc, outperforming the stronger of Uniform and Argmax by 1.5, 5.1, and 10.1 percentage points, respectively. These gains extend to other backbones. For example, on SVHalluc, PACD achieves 60.4\% with VideoLLaMA2-AV-7B, compared with 54.3\% for Uniform and 56.0\% for Argmax. These results demonstrate that simplified relevance weighting alone cannot match the benefits of PACD's candidate arbitration and consensus recovery.

The category-level results further highlight PACD's effectiveness across different task settings. On the visual-dominant CMM category, PACD achieves 86.5\% with VideoLLaMA2-AV-7B, surpassing Uniform (77.5\%) and Argmax (80.0\%). Similarly, on the semantic category of SVHalluc, PACD reaches 75.7\% with OmniVinci, substantially outperforming Uniform (57.0\%) and Argmax (59.6\%). These improvements further demonstrate the effectiveness of provenance-aware decoding in leveraging task-relevant modality evidence.

\begin{table*}[t]
    \centering
    \caption{\textbf{Complete component comparison of PACD.} Accuracy (\%) on both categories and the overall benchmark for CMM, AVHBench, and SVHalluc. Routing adopts $c_R$; RGCA adds provenance-guided candidate construction and confidence-based arbitration to obtain $c_B$; R+PBCR combines routing with consensus recovery while omitting that arbitration step. PACD combines RGCA and PBCR with the recovery priority defined in the main paper. Bold denotes the highest displayed result per model and metric, including ties after rounding.}
    \label{tab:components_ablation_full}

    \small
    \setlength{\tabcolsep}{0.5mm}
    \renewcommand{\arraystretch}{0.98}

    \begin{tabular*}{\textwidth}{
        @{\extracolsep{\fill}}
        p{1.30cm}
        C{1.70cm}
        C{1.70cm}
        C{1.70cm}
        C{2.05cm}
        C{2.05cm}
        C{2.05cm}
        C{1.05cm}
        C{1.05cm}
        C{2.10cm}
        @{}
    }
        \toprule

        \multirow[c]{2}{*}{\raisebox{-0.98em}[0pt][0pt]{\textbf{Method}}}
        & \multicolumn{3}{c}{\textbf{CMM}}
        & \multicolumn{3}{c}{\textbf{AVHBench}}
        & \multicolumn{3}{c}{\textbf{SVHalluc}} \\

        \cmidrule(lr){2-4}
        \cmidrule(lr){5-7}
        \cmidrule(lr){8-10}

        &
        \parbox[c][2.6em][c]{1.70cm}{
            \centering\textbf{Visual Dom.}
        }
        &
        \parbox[c][2.6em][c]{1.70cm}{
            \centering\textbf{Audio Dom.}
        }
        &
        \parbox[c][2.6em][c]{1.70cm}{
            \centering\textbf{Overall Acc.}
        }
        &
        \parbox[c][2.6em][c]{2.05cm}{
            \centering
            \textbf{Video-Driven}\\
            \textbf{Audio Hall.}
        }
        &
        \parbox[c][2.6em][c]{2.05cm}{
            \centering
            \textbf{Audio-Driven}\\
            \textbf{Video Hall.}
        }
        &
        \parbox[c][2.6em][c]{2.05cm}{
            \centering\textbf{Overall Acc.}
        }
        &
        \parbox[c][2.6em][c]{1.05cm}{
            \centering\textbf{Sem.}
        }
        &
        \parbox[c][2.6em][c]{1.05cm}{
            \centering\textbf{Temp.}
        }
        &
        \parbox[c][2.6em][c]{2.10cm}{
            \centering\mbox{\textbf{Overall Acc.}}
        }
        \\

        \midrule

        \multicolumn{10}{c}{\textbf{VideoLLaMA2-AV-7B}} \\
        \midrule

        Base
        & 72.3 & \textbf{82.3} & 77.3
        & 70.8 & 76.1 & 72.6
        & 57.2 & 43.8 & 51.7 \\

        Routing
        & 81.0 & 68.0 & 74.5
        & 78.2 & 76.5 & 77.6
        & 64.0 & 43.3 & 55.6 \\

        RGCA
        & 84.5 & 78.8 & 81.6
        & 81.1 & 79.0 & 80.4
        & 65.5 & 43.9 & 56.7 \\

        R+PBCR
        & 83.0 & 67.5 & 75.3
        & 78.4 & 76.6 & 77.8
        & \textbf{71.1} & 43.7 & 59.9 \\

        \rowcolor{pacdpurple}
        \textbf{PACD}
        & \textbf{86.5} & 78.3 & \textbf{82.4}
        & \textbf{81.2} & \textbf{79.4} & \textbf{80.6}
        & 71.0 & \textbf{45.0} & \textbf{60.4} \\

        \midrule
        \multicolumn{10}{c}{\textbf{OmniVinci}} \\
        \midrule

        Base
        & 81.0 & 83.8 & 82.4
        & 72.2 & 68.8 & 71.1
        & 54.5 & 45.8 & 50.9 \\

        Routing
        & 87.8 & 79.3 & 83.5
        & 77.7 & 67.6 & 74.3
        & 59.7 & 44.9 & 53.6 \\

        RGCA
        & 88.3 & 85.5 & 86.9
        & 81.6 & 69.8 & 77.7
        & 59.8 & 45.5 & 54.0 \\

        R+PBCR
        & 88.0 & 80.0 & 84.0
        & 78.7 & 69.3 & 75.6
        & 75.0 & 45.4 & 62.9 \\

        \rowcolor{pacdpurple}
        \textbf{PACD}
        & \textbf{88.5} & \textbf{86.3} & \textbf{87.4}
        & \textbf{82.6} & \textbf{71.0} & \textbf{78.8}
        & \textbf{75.7} & \textbf{46.4} & \textbf{63.7} \\

        \midrule
        \multicolumn{10}{c}{\textbf{Qwen2.5-Omni-3B}} \\
        \midrule

        Base
        & 70.8 & 85.3 & 78.0
        & 73.4 & 80.5 & 75.8
        & 72.7 & 42.1 & 60.2 \\

        Routing
        & \textbf{81.0} & 85.0 & 83.0
        & 81.9 & 79.2 & 81.0
        & 72.9 & 45.1 & 61.5 \\

        RGCA
        & 80.8 & 86.0 & 83.4
        & 82.6 & 81.6 & 82.3
        & \textbf{78.2} & 45.8 & 64.9 \\

        R+PBCR
        & 80.5 & 87.8 & 84.1
        & 81.7 & 80.3 & 81.3
        & 73.1 & 46.8 & 62.4 \\

        \rowcolor{pacdpurple}
        \textbf{PACD}
        & 80.8 & \textbf{88.5} & \textbf{84.6}
        & \textbf{82.7} & \textbf{81.8} & \textbf{82.4}
        & 78.0 & \textbf{47.6} & \textbf{65.6} \\

        \midrule
        \multicolumn{10}{c}{\textbf{Qwen2.5-Omni-7B}} \\
        \midrule

        Base
        & 66.5 & 70.3 & 68.4
        & 73.0 & 80.4 & 75.5
        & 77.5 & 46.2 & 64.7 \\

        Routing
        & 84.8 & 88.3 & 86.5
        & 80.0 & 84.4 & 81.4
        & 76.1 & 45.2 & 63.5 \\

        RGCA
        & 84.8 & 88.3 & 86.5
        & 80.6 & \textbf{84.7} & 81.9
        & \textbf{83.0} & 46.1 & 67.9 \\

        R+PBCR
        & \textbf{85.0} & \textbf{89.5} & \textbf{87.3}
        & 80.0 & 84.4 & 81.5
        & 75.9 & 47.4 & 64.2 \\

        \rowcolor{pacdpurple}
        \textbf{PACD}
        & \textbf{85.0} & \textbf{89.5} & \textbf{87.3}
        & \textbf{80.7} & \textbf{84.7} & \textbf{82.0}
        & 82.8 & \textbf{48.5} & \textbf{68.8} \\

        \bottomrule
    \end{tabular*}
\end{table*}

\begin{table*}[t]
    \centering
    \caption{\textbf{Task-level component comparison on SVHalluc.} Routing, RGCA, R+PBCR, and PACD follow the definitions in Sec.~\ref{sec:supp_components}. Task abbreviations follow Table~\ref{tab:svhalluc_detailed}. Avg. is question-weighted accuracy over all 2,405 questions. All values are percentages. Bold denotes the highest displayed result within each backbone and column, including ties after rounding.}
    \label{tab:svhalluc_detailed_component_ablation}
    \small
    \setlength{\tabcolsep}{3pt}
    \renewcommand{\arraystretch}{1.02}
    \begin{tabularx}{\textwidth}{@{}>{\raggedright\arraybackslash}p{1.5cm}*{7}{>{\centering\arraybackslash}X}@{}}
        \toprule
        \multirow{2}{*}{\textbf{Method}} & \multicolumn{3}{c}{\textbf{Sem.}} & \multicolumn{3}{c}{\textbf{Temp.}} & \multirow{2}{*}{\textbf{Avg.}} \\
        \cmidrule(lr){2-4}\cmidrule(lr){5-7}
        & \textbf{GSA} & \textbf{FGSA} & \textbf{CMSB} & \textbf{TA} & \textbf{TF} & \textbf{CMTB} & \\
        \midrule
        \multicolumn{8}{c}{\textbf{VideoLLaMA2-AV-7B}} \\
        \midrule
        Base & \textbf{50.0} & 57.2 & 63.1 & \textbf{50.0} & 26.8 & \textbf{51.0} & 51.7 \\
        Routing & \textbf{50.0} & 60.1 & 78.8 & \textbf{50.0} & 31.2 & 46.0 & 55.6 \\
        RGCA & \textbf{50.0} & 70.0 & 74.7 & \textbf{50.0} & 30.1 & 48.7 & 56.7 \\
        R+PBCR & \textbf{50.0} & 81.5 & \textbf{80.0} & \textbf{50.0} & 32.2 & 46.3 & 59.9 \\
        \rowcolor{pacdpurple}
        \textbf{PACD} & \textbf{50.0} & \textbf{82.4} & 79.0 & \textbf{50.0} & \textbf{32.6} & 49.6 & \textbf{60.4} \\
        \midrule
        \multicolumn{8}{c}{\textbf{OmniVinci}} \\
        \midrule
        Base & 50.0 & 51.1 & 61.0 & 50.0 & \textbf{34.1} & 50.7 & 50.9 \\
        Routing & 50.0 & 51.8 & 74.3 & 50.0 & 33.3 & 48.7 & 53.6 \\
        RGCA & 50.0 & 55.2 & 71.9 & 50.0 & 33.3 & 50.4 & 54.0 \\
        R+PBCR & 67.6 & \textbf{80.0} & 77.2 & 51.1 & 33.3 & 49.0 & 62.9 \\
        \rowcolor{pacdpurple}
        \textbf{PACD} & \textbf{68.7} & 79.5 & \textbf{78.3} & \textbf{51.9} & 33.3 & \textbf{51.0} & \textbf{63.7} \\
        \midrule
        \multicolumn{8}{c}{\textbf{Qwen2.5-Omni-3B}} \\
        \midrule
        Base & \textbf{72.1} & 72.3 & 73.6 & 49.2 & 22.8 & \textbf{50.1} & 60.2 \\
        Routing & 50.9 & 82.0 & 83.7 & 50.0 & 33.3 & 49.3 & 61.5 \\
        RGCA & 64.6 & \textbf{84.5} & \textbf{84.3} & 51.1 & 35.1 & 48.7 & 64.9 \\
        R+PBCR & 51.6 & 82.0 & 83.7 & 49.2 & 39.5 & \textbf{50.1} & 62.4 \\
        \rowcolor{pacdpurple}
        \textbf{PACD} & 64.6 & 84.2 & 83.9 & \textbf{51.4} & \textbf{39.9} & 49.9 & \textbf{65.6} \\
        \midrule
        \multicolumn{8}{c}{\textbf{Qwen2.5-Omni-7B}} \\
        \midrule
        Base & \textbf{72.5} & 84.0 & 76.2 & 52.7 & 33.3 & 49.6 & 64.7 \\
        Routing & 49.5 & 89.4 & 87.1 & 50.3 & 33.3 & 49.3 & 63.5 \\
        RGCA & 71.2 & 89.2 & \textbf{87.6} & \textbf{53.8} & 33.3 & 48.1 & 67.9 \\
        R+PBCR & 49.1 & \textbf{90.1} & 86.3 & 50.3 & \textbf{38.4} & \textbf{51.6} & 64.2 \\
        \rowcolor{pacdpurple}
        \textbf{PACD} & 70.7 & 89.9 & 86.9 & \textbf{53.8} & \textbf{38.4} & 51.0 & \textbf{68.8} \\
        \bottomrule
    \end{tabularx}
\end{table*}

\subsection{Fine-Grained Results on SVHalluc}
\label{sec:supp_svhalluc}

Tables~\ref{tab:svhalluc_detailed} and~\ref{tab:svhalluc_detailed_fusion_ablation} provide the six-task breakdown for the main baselines and the simplified decoding strategies, respectively. GSA, FGSA, and CMSB belong to the semantic category, while TA, TF, and CMTB belong to the temporal category.

FGSA exhibits improvements over MAD across all four backbones. PACD increases FGSA accuracy from 69.6\% to 82.4\% on VideoLLaMA2-AV-7B and from 55.4\% to 79.5\% on OmniVinci; the corresponding values are 81.8\% to 84.2\% on Qwen2.5-Omni-3B and 89.2\% to 89.9\% on Qwen2.5-Omni-7B. OmniVinci also improves from 50.0\% to 68.7\% on GSA and from 70.0\% to 78.3\% on CMSB. These task-level changes help explain its higher semantic accuracy in the aggregate comparison.

PACD also demonstrates improvements on temporal reasoning tasks. On TF, PACD increases accuracy from 32.2\% to 39.9\% with Qwen2.5-Omni-3B and from 33.3\% to 38.4\% with Qwen2.5-Omni-7B compared with MAD. On CMTB, Qwen2.5-Omni-7B further improves from 50.4\% to 51.0\%. Together with the improvements on semantic tasks, these results highlight PACD's effectiveness across different aspects of audio-visual understanding and contribute to its superior overall performance on SVHalluc.

\section{Component Analysis}
\label{sec:supp_components}

Table~\ref{tab:components_ablation_full} extends the main-paper component comparison to all four models. The variants correspond to the following prediction procedures.
\begin{itemize}
\item \textbf{Routing} selects $m^*$ from $r_A$ and $r_V$ using the main-paper tie rule and adopts the corresponding branch prediction $c_R$.
\item \textbf{RGCA} additionally constructs the alternative $c^*$ from provenance evidence and applies confidence-based arbitration, producing $c_B$. Consensus recovery is disabled.
\item \textbf{R+PBCR} retains routing and the provenance computations needed to identify $c^*$, but omits confidence-based arbitration. Its non-recovery prediction is $c_R$; when the PBCR conditions hold, it selects $c^*$.
\item \textbf{PACD} combines Relevance Guided Candidate Arbitration (RGCA) with Provenance Based Consensus Recovery (PBCR). For non-consensus cases, it outputs $c_B$. For consensus cases, it selects $c^*$ when the recovery condition holds and otherwise retains $c_{\mathrm{MAD}}$.
\end{itemize}

The ablation results highlight the complementary contributions of RGCA and PBCR. On OmniVinci, RGCA improves overall accuracy over Routing from 83.5\% to 86.9\% on CMM and from 74.3\% to 77.7\% on AVHBench. On SVHalluc, R+PBCR substantially improves accuracy from 53.6\% to 62.9\%, while the complete PACD further reaches 63.7\%. Similar improvements are observed with VideoLLaMA2-AV-7B, where Routing, RGCA, and R+PBCR achieve 55.6\%, 56.7\%, and 59.9\%, respectively, compared with 60.4\% for PACD. These results demonstrate the complementary benefits of preserving reliable modality-specific predictions through RGCA and recovering incorrect consensus through PBCR.

Table~\ref{tab:svhalluc_detailed_component_ablation} further illustrates the contributions of each component across individual SVHalluc tasks. On OmniVinci, R+PBCR improves FGSA accuracy from 51.8\% under Routing to 80.0\%, while PACD achieves improvements on GSA, CMSB, TA, and CMTB compared with R+PBCR. On Qwen2.5-Omni-7B, PACD achieves the highest overall accuracy of 68.8\%, surpassing RGCA (67.9\%) and R+PBCR (64.2\%). These findings further demonstrate that integrating both components yields stronger overall performance across diverse semantic and temporal reasoning tasks.

\section{Performance across Duration Groups}
\label{sec:supp_duration}

\begin{figure*}[t]
    \centering
    \includegraphics[width=\textwidth]{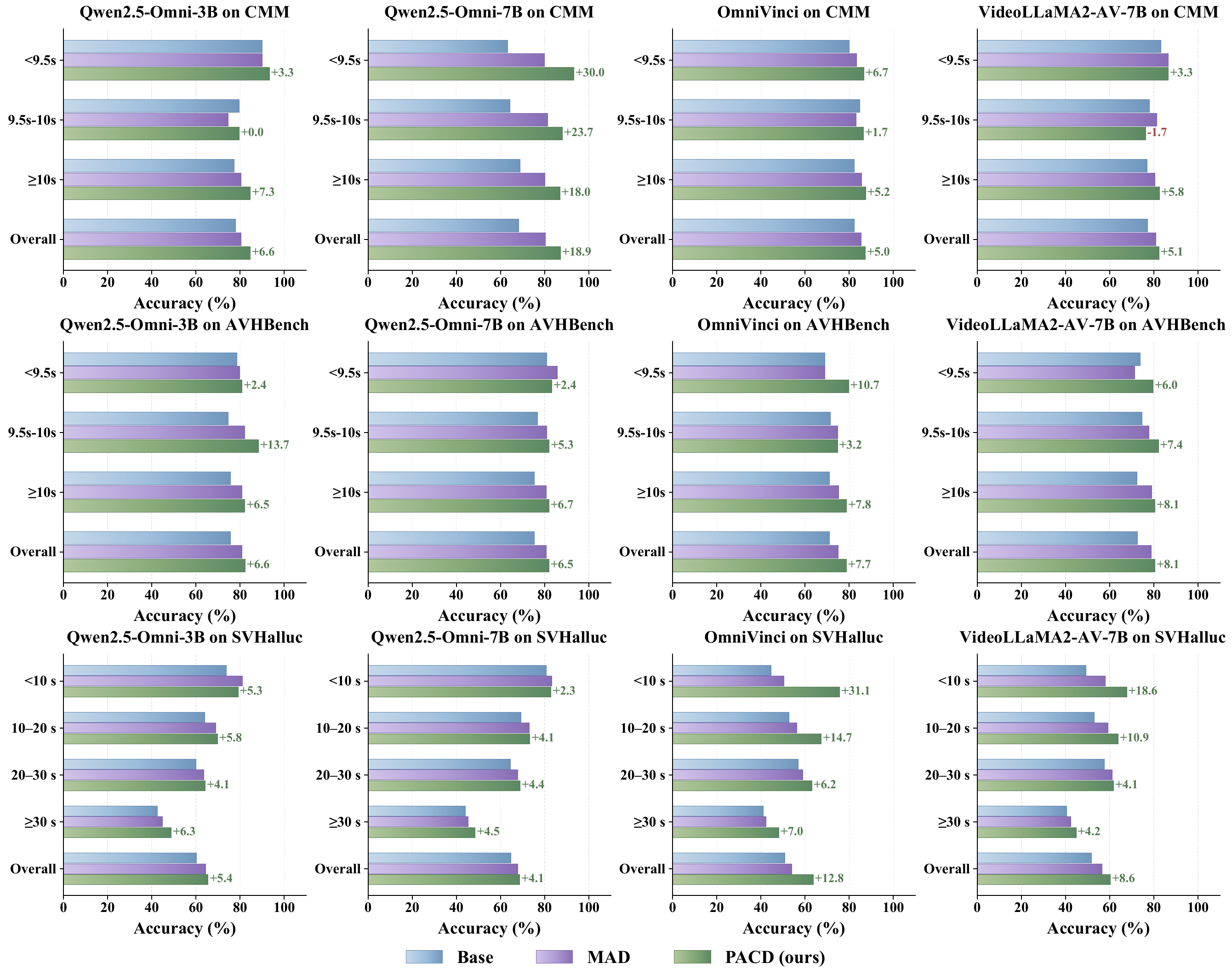}
    \caption{\textbf{Accuracy across prepared-input duration groups.} Rows correspond to CMM, AVHBench, and SVHalluc, and columns correspond to the four backbones. Blue, purple, and green bars show Base, MAD, and PACD, respectively. Annotations next to PACD bars give its accuracy change from Base in percentage points.}
    \label{fig:duration}
\end{figure*}

Figure~\ref{fig:duration} compares the accuracy of Base, MAD, and PACD across different video duration groups. Videos are grouped by the duration of their prepared 2-fps inputs, using consistent duration boundaries across all four backbones.

For CMM and AVHBench, videos are divided into three duration groups: $d<9.5$\,s, $9.5\leq d<10$\,s, and $d\geq10$\,s. SVHalluc uses four groups: $d<10$\,s, $10\leq d<20$\,s, $20\leq d<30$\,s, and $d\geq30$\,s. Overall accuracy is calculated across all questions in each benchmark.

PACD consistently improves accuracy across different video duration groups. On SVHalluc, OmniVinci achieves a substantial gain of 31.1 percentage points over Base in the shortest duration group, with an overall improvement of 12.8 points. Similarly, on CMM, Qwen2.5-Omni-7B improves by 30.0 points in the shortest group and 18.9 points overall. These results highlight PACD's effectiveness across varying video durations and different Omni-LLM backbones.

\section{Candidate-Score Discrimination and Decision Operating Points}
\label{sec:score_roc}

\begin{figure*}[t]
    \centering
    \includegraphics[width=\textwidth]{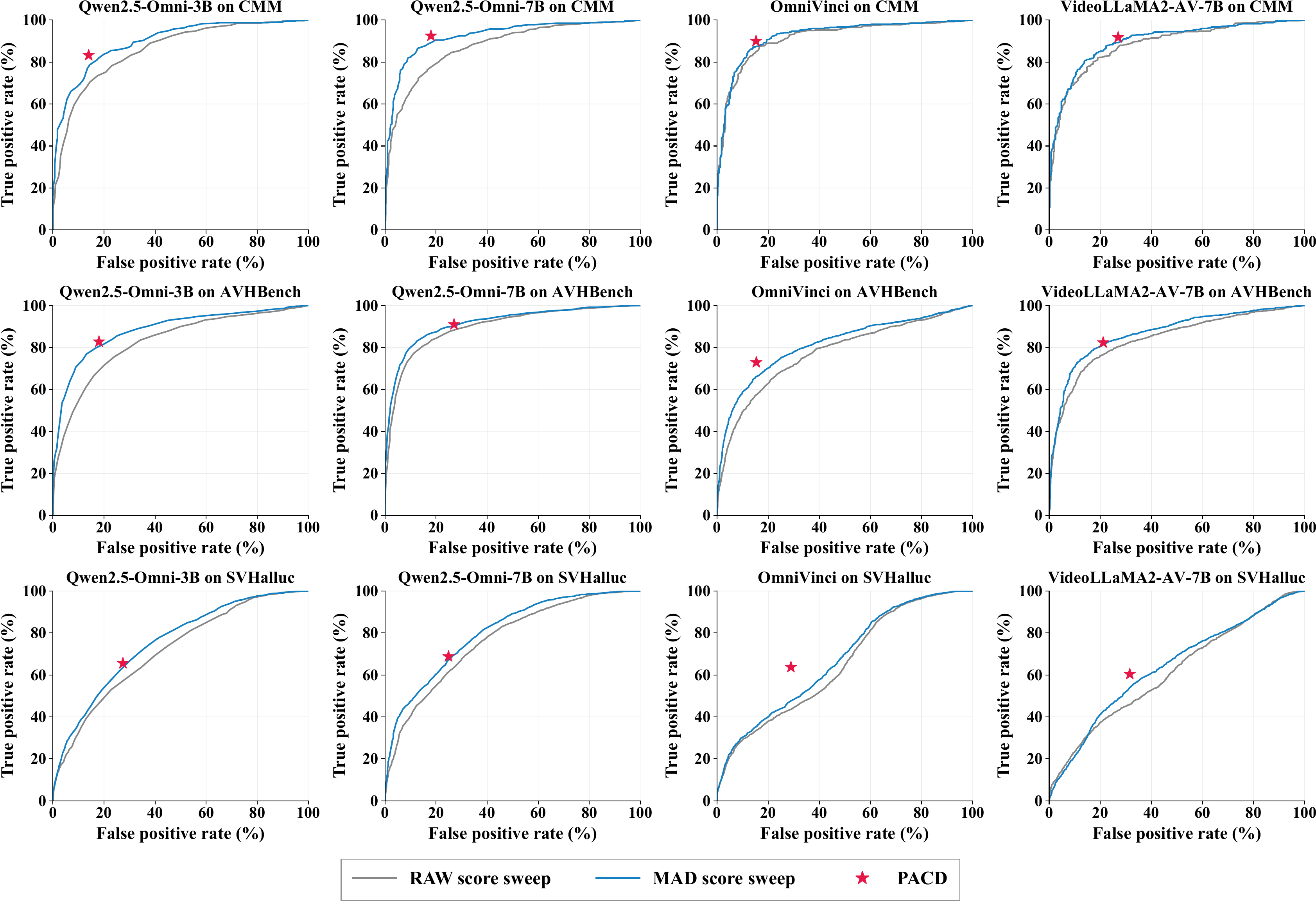}
    \caption{\textbf{Candidate-score discrimination and PACD operating points.} Gray and blue curves represent ROC threshold sweeps of candidate-logit margins from the joint audio-visual branch (RAW) and MAD, respectively. Red stars indicate the operating points of PACD. Columns correspond to four Omni-LLMs, and rows correspond to CMM, AVHBench, and SVHalluc. CMM and AVHBench use binary ROC with \textit{Yes} as the positive class, while SVHalluc uses pooled candidate-level one-versus-rest micro-ROC over two- and three-choice questions.}
    \label{fig:score_roc}
\end{figure*}

We investigate whether the decision performance of PACD can be achieved by simply adjusting the candidate-score thresholds of the joint audio-visual branch or MAD. Figure~\ref{fig:score_roc} compares their ROC curves with the operating points of PACD across four Omni-LLMs and three benchmarks. Following the main paper, the valid candidate set for question $i$ is $\mathcal{C}_i=\{c_{i,1},\ldots,c_{i,K_i}\}$, with candidate logit $z_b^t(c)=z_b^t(\phi_i(c))$. The gray RAW curve corresponds to the joint audio-visual branch ($b=AV$), while the blue curve represents MAD ($b=\mathrm{MAD}$). We denote the final PACD prediction by $\hat{c}_{\mathrm{PACD},i}$ and the ground-truth answer by $c_i^{\mathrm{gt}}$.

\paragraph{Binary benchmarks.} For CMM and AVHBench, we treat \textit{Yes} as the positive class and vary a decision threshold $\eta$ according to
\begin{equation}
    \text{predict \textit{Yes} if } z_b^t(\mathrm{Yes})-z_b^t(\mathrm{No})>\eta,
    \label{eq:supp_binary_sweep}
\end{equation}
with \textit{No} predicted otherwise. Samples with identical score margins are processed together to obtain the corresponding true-positive rates (TPR) and false-positive rates (FPR). The PACD operating point is directly determined by its final predictions $\mathbb{I}[\hat{c}_{\mathrm{PACD},i}=\mathrm{Yes}]$.

\paragraph{Mixed-candidate benchmark.} SVHalluc contains both two- and three-choice questions ($K_i\in\{2,3\}$). For each valid candidate, we compute its margin against the strongest competing candidate and apply the threshold $\eta$:
\begin{equation}
    \mathbb{I}\!\left[ z_b^t(c)-\max_{c'\in\mathcal{C}_i\setminus\{c\}}z_b^t(c')>\eta \right].
    \label{eq:supp_candidate_sweep}
\end{equation}
Using $\mathbb{I}[c=c_i^{\mathrm{gt}}]$ as the ground-truth label, we construct a pooled candidate-level one-versus-rest micro-ROC over 2,405 positive and 3,020 negative question--candidate pairs. The corresponding PACD decision is $\mathbb{I}[c=\hat{c}_{\mathrm{PACD},i}]$. Since PACD selects exactly one candidate per question, its TPR equals question-level accuracy, while its FPR is the number of incorrect predictions divided by 3,020. Unlike PACD, threshold-based candidate selection may retain zero or multiple candidates for a given question. Therefore, comparisons are conducted within each benchmark using its corresponding evaluation protocol.

\paragraph{Comparison at matched false-positive rates.} MAD achieves higher candidate-score AUROC than the joint audio-visual branch across all 12 model--benchmark combinations. To further assess PACD, we compare its operating point with the highest TPR attainable by MAD under the same or lower FPR, using the observed deterministic threshold sweeps.

PACD consistently achieves a higher TPR than the corresponding MAD threshold sweep across all 12 combinations. On CMM with Qwen2.5-Omni-3B, PACD achieves a TPR of 83.25\% at an FPR of 14.00\%, whereas MAD reaches at most 77.00\% under the same FPR constraint. Similarly, on SVHalluc with OmniVinci, PACD achieves a TPR of 63.70\% at an FPR of 28.91\%, substantially exceeding MAD's maximum of 47.53\%. These results demonstrate that PACD provides more favorable decision trade-offs than those attainable through global threshold adjustment of MAD's candidate scores.

\section{Stability under Media-Level Resampling}
\label{sec:supp_resampling}

We further evaluate the stability of PACD's accuracy improvements using a paired media-cluster bootstrap. This approach accounts for dependencies among questions associated with the same media while preserving paired comparisons between PACD and Base.

For each benchmark, we randomly sample $G$ media IDs with replacement, where $G$ denotes the number of distinct media groups, and include all associated questions in each replicate. Identical samples are used to evaluate PACD and Base. CMM contains 800 singleton groups, while AVHBench and SVHalluc contain 1,287 and 872 media groups, respectively. Accordingly, the procedure reduces to a paired question-level bootstrap for CMM.

Let $\mathcal{I}_g$ denote the set of questions associated with media group $g$, and $\nu_g^{(b)}$ its sampling multiplicity in replicate $b$. We define $d_i$ as the difference between the correctness indicators of PACD and Base for question $i$, where $\hat{c}_{\mathrm{Base},i}=c_{AV,i}$. The accuracy improvement for each replicate is calculated as
\begin{equation}
    \Delta^{(b)}=100\frac{\displaystyle\sum_{g=1}^{G}\nu_g^{(b)}\sum_{i\in\mathcal{I}_g}d_i}{\displaystyle\sum_{g=1}^{G}\nu_g^{(b)}|\mathcal{I}_g|}.
    \label{eq:supp_bootstrap_gain}
\end{equation}
We perform 5,000 bootstrap replicates and compute the 95\% percentile intervals using the 2.5th and 97.5th percentiles of the resulting accuracy improvements.

\begin{figure*}[t]
    \centering
    \includegraphics[width=\textwidth]{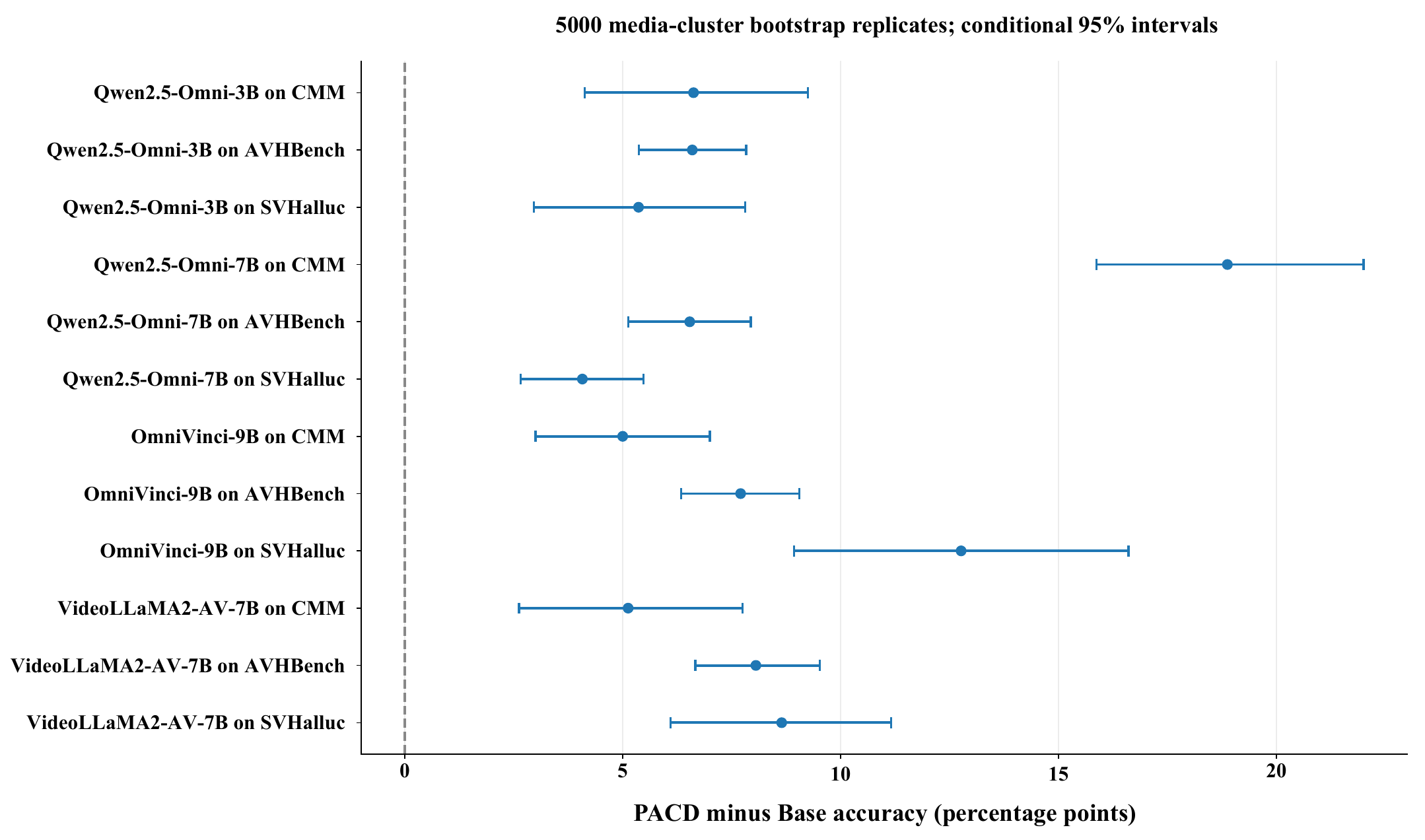}
    \caption{\textbf{Stability of PACD's accuracy improvements under media-level resampling.} Dots indicate the accuracy gains of PACD over Base in percentage points, while horizontal bars represent 95\% percentile intervals obtained from 5,000 paired media-cluster bootstrap replicates. Media IDs are sampled with replacement, with identical samples used for both methods. The dashed vertical line denotes zero improvement. All intervals lie above zero across the 12 model--benchmark combinations.}
    \label{fig:media}
\end{figure*}

As shown in Fig.~\ref{fig:media}, PACD consistently outperforms Base across all 12 model--benchmark combinations, with accuracy improvements ranging from 4.07 to 18.88 percentage points. Notably, all 95\% bootstrap intervals lie entirely above zero, indicating stable improvements under media-level resampling. On CMM, Qwen2.5-Omni-7B achieves a gain of 18.88 points, with an interval of $[15.88,22.00]$. On SVHalluc, OmniVinci improves by 12.77 points with an interval of $[8.93,16.61]$, while VideoLLaMA2-AV-7B gains 8.65 points with an interval of $[6.10,11.16]$. These results further demonstrate the consistency and robustness of PACD's performance improvements across different Omni-LLMs and benchmarks.

\section{Qualitative Examples}
\label{sec:supp_qualitative}

\begin{figure*}[t]
    \centering
    \includegraphics[width=\textwidth]{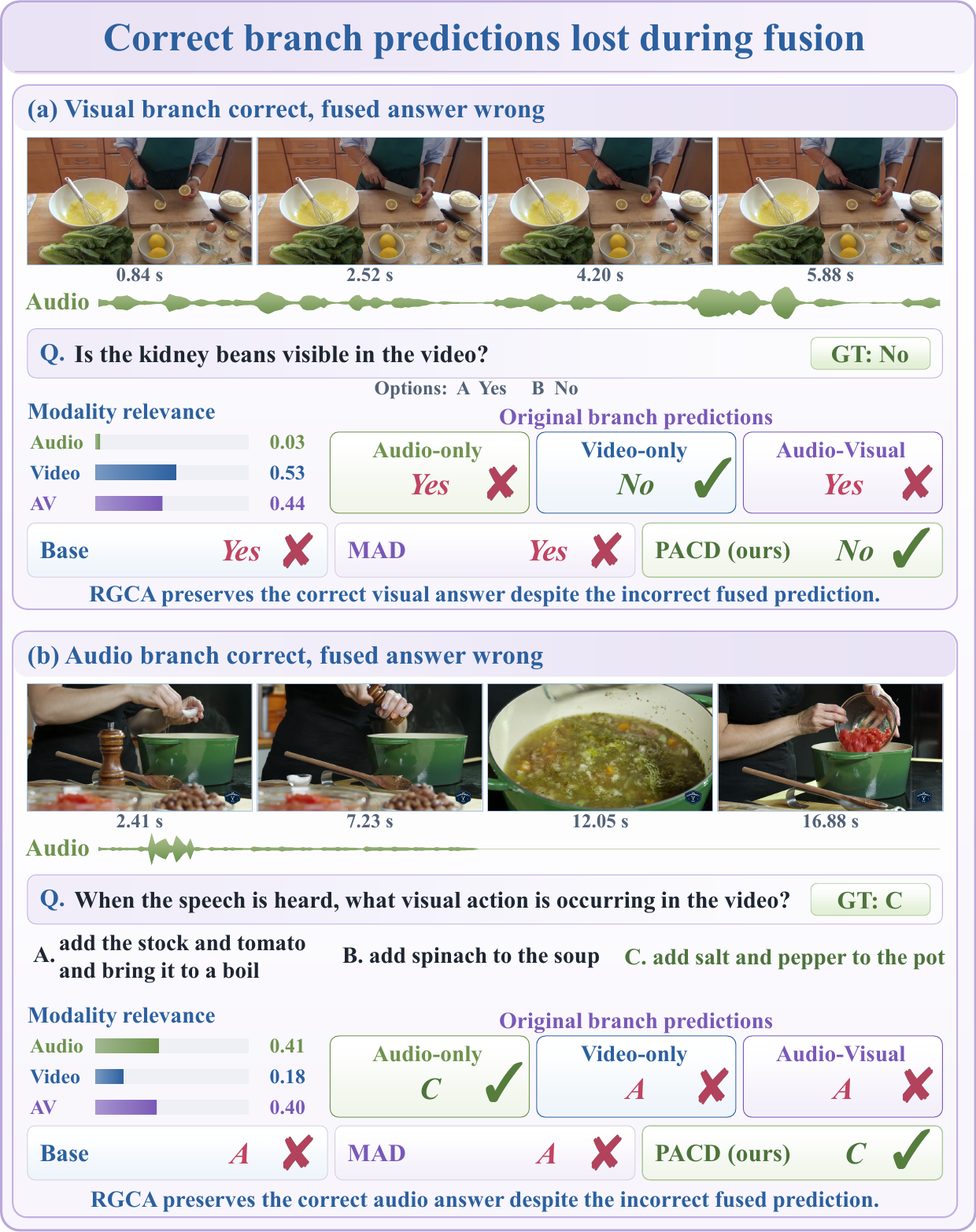}
    \caption{Qualitative examples of PACD preserving correct modality-specific predictions during fusion.}
    \label{fig:qualitative_fusion}
\end{figure*}

\begin{figure*}[t]
    \centering
    \includegraphics[width=\textwidth]{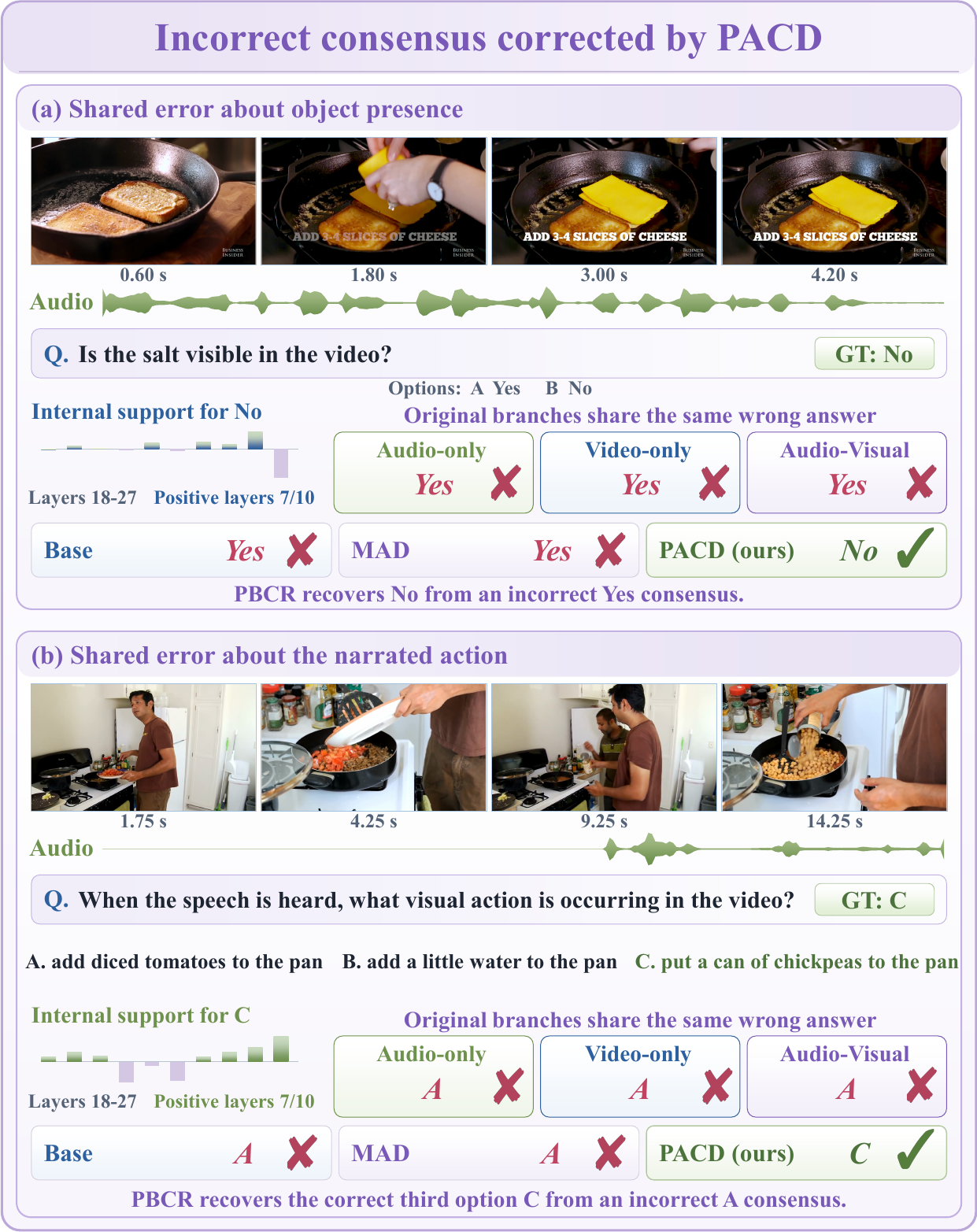}
    \caption{Qualitative examples of PACD correcting incorrect consensus across audio-only, video-only, and audio-visual branches.}
    \label{fig:qualitative_consensus}
\end{figure*}

We present qualitative examples of PACD on OmniVinci with SVHalluc to further illustrate how our method addresses the two challenges identified in the main-paper motivation: correct modality-specific predictions being lost during fusion and incorrect consensus among modality-specific branches. Figures~\ref{fig:qualitative_fusion} and~\ref{fig:qualitative_consensus} show how PACD produces correct predictions in both scenarios.

\paragraph{Preserving correct predictions during fusion.} Figure~\ref{fig:qualitative_fusion} presents two examples where the relevant unimodal branch predicts the correct answer, but the joint audio-visual prediction is incorrect. In Fig.~\ref{fig:qualitative_fusion}(a), the question asks whether kidney beans are visible in the video. The video-only branch correctly predicts \textit{No}, while the audio-only and audio-visual branches predict \textit{Yes}. PACD successfully preserves the correct visual prediction, consistent with the higher relevance assigned to video ($r_V=0.53$). In Fig.~\ref{fig:qualitative_fusion}(b), the question asks about the visual action occurring when speech is heard. The audio-only branch correctly selects option C, whereas the video-only and audio-visual branches select option A. PACD again produces the correct answer by retaining the prediction supported by the relevant unimodal branch. These examples demonstrate how PACD effectively preserves reliable modality-specific predictions that are otherwise lost during audio-visual fusion.

\paragraph{Correcting incorrect consensus.} Figure~\ref{fig:qualitative_consensus} illustrates two examples where all three modality-specific and joint branches agree on the same incorrect answer. In Fig.~\ref{fig:qualitative_consensus}(a), the audio-only, video-only, and audio-visual branches all predict \textit{Yes} when asked whether salt is visible in the video, while the ground-truth answer is \textit{No}. Despite this incorrect consensus, PACD identifies support for the alternative answer from layer-wise provenance evidence, with positive contributions across 7 of the 10 examined layers, and correctly predicts \textit{No}. Similarly, in Fig.~\ref{fig:qualitative_consensus}(b), all three branches incorrectly select option A for a question about the visual action occurring during speech. PACD leverages the internal evidence supporting option C, again with positive contributions in 7 of the 10 examined layers, to recover the correct answer. These examples highlight PACD's ability to correct hallucinations even when all modality-conditioned branches produce the same incorrect prediction.

\end{document}